\documentclass[10pt,twocolumn,letterpaper]{article}

\usepackage{comment}
\usepackage{amsmath,bm}
\usepackage{makecell} 
\usepackage{multirow}
\usepackage{booktabs}
\usepackage[accsupp]{axessibility}  % Improves PDF readability for those with disabilities.
\usepackage[pagenumbers]{cvpr} % To force page numbers, e.g. for an arXiv version

\definecolor{cvprblue}{rgb}{0.21,0.49,0.74}
\usepackage[pagebackref,breaklinks,colorlinks,allcolors=cvprblue]{hyperref}

\def\paperID{3135} % *** Enter the Paper ID here
\def\confName{CVPR}
\def\confYear{2026}

\title{Differentiable Vector Quantization for Rate-Distortion Optimization of Generative Image Compression}

\author{Shiyin Jiang$^{1}$ \quad Wei Long$^{1}$ \quad Minghao Han$^1$ \quad Zhenghao Chen$^2$ \quad Ce Zhu$^{1}$ \quad Shuhang Gu$^{1}$\footnotemark[1]\\
$^1$University of Electronic Science and Technology of China \quad $^2$The University of Newcastle, Australia\\
{\tt \small \{shiyin.jsy, shuhanggu\}@gmail.com} \\
}

\begin{document}
\maketitle

\renewcommand{\thefootnote}{\fnsymbol{footnote}}
\footnotetext[1]{corresponding author}

\begin{abstract}
The rapid growth of visual data under stringent storage and bandwidth constraints makes \emph{extremely low-bitrate} image compression increasingly important. While Vector Quantization (VQ) offers strong structural fidelity, existing methods lack a principled mechanism for joint rate--distortion (RD) optimization due to the disconnect between representation learning and entropy modeling.
We propose RDVQ, a unified framework that enables end-to-end RD optimization for VQ-based compression via a differentiable relaxation of the codebook distribution, allowing the entropy loss to directly shape the latent prior. We further develop an autoregressive entropy model that supports accurate entropy modeling and test-time rate control.
Extensive experiments demonstrate that RDVQ achieves strong performance at extremely low bitrates with a lightweight architecture, attaining competitive or superior perceptual quality with significantly fewer parameters. Compared with RDEIC, RDVQ reduces bitrate by up to 75.71\% on DISTS and 37.63\% on LPIPS on DIV2K-val. Beyond empirical gains, RDVQ introduces an entropy-constrained formulation of VQ, highlighting the potential for a more unified view of image tokenization and compression. The code will be available at \url{https://github.com/CVL-UESTC/RDVQ}.
\end{abstract}
    
\begin{figure}[htbp]     
  \centering
  \includegraphics[width=0.99\columnwidth]{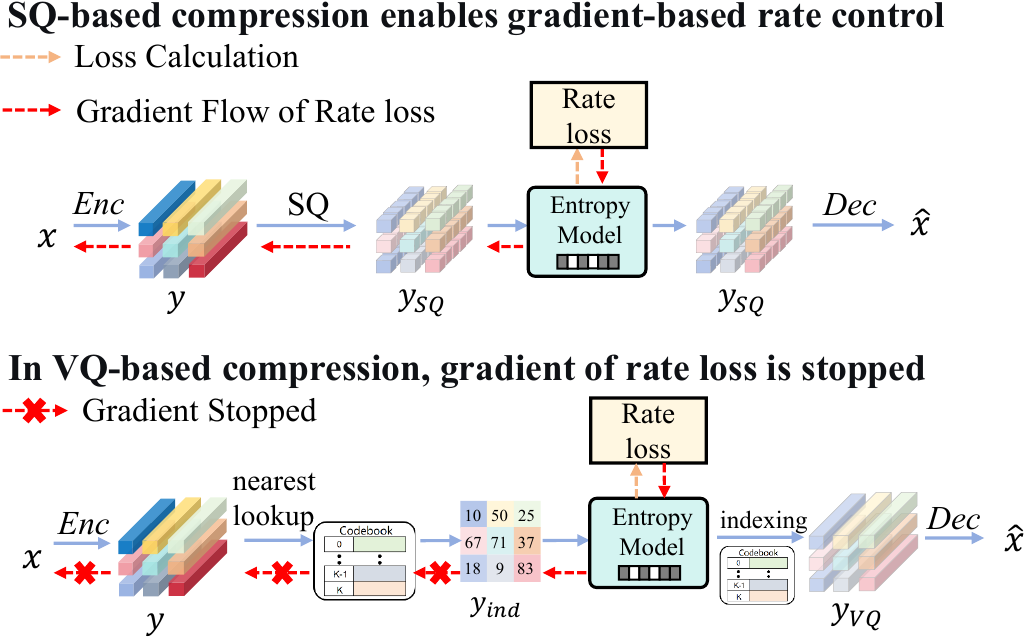} % 或设为0.9\columnwidth 等
  \vspace{-0.2cm}
  \caption{Gradient-based rate control in VQ compression is challenging due to non-differentiable indices.}
  \label{fig:Motivation}
\end{figure}

\section{Introduction}
\label{sec:intro}
The rapid growth of visual data has intensified the demand for high-quality lossy compression under stringent storage and bandwidth constraints. Conventional standards~\cite{BPG,VVC}, constrained by hand-crafted designs, struggle to achieve optimal coding efficiency. In contrast, learned image compression optimizes the rate--distortion (RD) objective in an end-to-end manner, yielding superior RD trade-offs~\cite{balle2018variational}. To further enhance perceptual quality at low bitrates, Generative Image Compression (GIC)~\cite{HiFiC, HiFiSD} incorporates generative priors, enabling the reconstruction of visually plausible details beyond pixel-wise fidelity.

Most learned codecs, including GIC models, follow a joint RD optimization framework~\cite{balle2018variational}, typically consisting of a transform autoencoder, a quantizer, and a learned entropy model. The coding rate is minimized via the cross-entropy $R=\text{CE}(p, q)$ between the marginal distribution of quantized latents $p$ (induced by the encoder) and the entropy model prediction $q$. Effective RD optimization therefore requires both distortion and rate terms to jointly influence the encoder, aligning representation learning with entropy modeling.
In practice, scalar quantization (SQ) enables such joint optimization through differentiable approximations (\eg, additive noise or the Straight-Through Estimator). However, its element-wise formulation fails to capture cross-channel dependencies, often leading to structural degradation under aggressive compression. Vector Quantization (VQ), by contrast, maps features to discrete codebook atoms that encode joint structural and semantic patterns, offering improved perceptual fidelity~\cite{vqkmeans, xue2025dlf,lu2024hybridflow,li2025onceforall}.

However, existing VQ-based methods typically adopt an implicit or uniform prior over codebook indices, as no explicit entropy constraint is imposed during training. While this design improves representation quality, it lacks a principled mechanism for entropy minimization. More critically, the discrete nearest-neighbor assignment prevents gradients from the rate objective from propagating back to the encoder. Consequently, the encoder-induced latent distribution (\ie, the prior $p$) remains largely unconstrained, and the entropy model can only passively fit this distribution without influencing it. This leads to a fundamental decoupling between representation learning and entropy modeling, hindering true end-to-end RD optimization.
To mitigate this limitation, prior works rely on indirect heuristics for bitrate control, such as adjusting codebook sizes~\cite{vqkmeans}, selective transmission~\cite{li2025onceforall, uigc}, or uniform coding schemes~\cite{xue2025dlf,lu2024hybridflow}. Although some approaches introduce learned entropy models~\cite{VQEntropy2, uigc}, they still lack a differentiable connection between the rate objective and the encoder, leaving the underlying prior effectively fixed and suboptimal.

In this paper, we propose \textbf{RDVQ}, a unified framework for end-to-end rate--distortion optimization in VQ-based compression. The core idea is to replace hard nearest-neighbor assignments with a differentiable, distance-aware soft distribution over codebook entries. This relaxation enables gradients from the rate objective to propagate to the encoder, allowing the entropy loss to jointly optimize both the encoder-induced latent distribution ($p$) and its entropy model ($q$). As a result, the traditionally implicit or uniform VQ prior becomes a learnable, entropy-aware prior that is fully integrated into the RD optimization process.
Building on this formulation, we further develop a masked-transformer-based autoregressive entropy model operating on the relaxed distribution. This model not only provides accurate entropy estimation for rate optimization and entropy coding, but also serves as a generative predictor, enabling test-time rate control within a limited operating range via prefix transmission and completion, without retraining.

RDVQ achieves strong performance at extremely low bitrates with a lightweight architecture. Without relying on large pretrained backbones, it attains competitive or superior perceptual quality using standard objectives such as GAN~\cite{GAN} and LPIPS~\cite{lpips}, while using less than $20\%$ of the parameters of prior methods. Compared with RDEIC~\cite{li2025rdeic}, RDVQ reduces bitrate by up to 75.71\% on DISTS~\cite{DISTS} and 37.63\% on LPIPS~\cite{lpips} on the DIV2K-val~\cite{Div2k} dataset.
Beyond empirical gains, RDVQ introduces an entropy-constrained formulation of VQ that connects representation learning and compression. This formulation suggests broader applicability: existing VQ-based image tokenizers, typically trained without entropy constraints, may be adapted into compression models under our framework, while incorporating entropy-aware learning may also improve the efficiency and structure of visual tokenizers. Together, these observations point toward a more unified view of image tokenization and compression.

Our contributions are as follows:
\begin{itemize}

\item We present \textbf{RDVQ}, a principled framework for \emph{end-to-end joint rate--distortion optimization in VQ-based compression}. To enable this, we introduce a differentiable relaxation of the discrete index distribution, which allows gradients from the rate objective to propagate to the encoder. This mechanism enables the entropy loss to directly shape the encoder-induced latent distribution, transforming the traditionally uniform VQ prior into a \emph{learnable and entropy-aware prior}.

\item Building upon this formulation, we develop a transformer-based autoregressive entropy model that operates on the relaxed distribution. It provides accurate entropy estimation for rate optimization and coding, while also serving as a \emph{generative predictor}, enabling \emph{test-time rate control within a limited operating range} via prefix transmission and completion without retraining.

\item We demonstrate that RDVQ achieves strong performance at low bitrates with a lightweight architecture, achieving competitive or superior perceptual quality with significantly fewer parameters. Beyond empirical results, RDVQ introduces an entropy-constrained VQ formulation connecting representation learning and compression.

\end{itemize}
\section{Related work}
\label{sec:rela}

\subsection{SQ \textit{\textbf{vs}} VQ in image compression}

In transform coding–based lossy image compression~\cite{balle2018variational,liu2023tcm,HeELiC,minnen2020channel,BalleEnd,cca,li2023frequency,sun2020end}, quantization converts continuous latent features into discrete symbols for entropy coding, thereby critically shaping the rate–distortion (RD) trade-off.

Most existing methods adopt Scalar Quantization (SQ)~\cite{xie2021enhanced,zou2022devil,liu2023img2,liu2025img3,jiang2023mlic++,li2025rdeic,mentzer2023m2t,lu2025learned,cheng2020anchor}, which quantizes each latent element independently (\eg, via rounding) or formulates quantization as a classification task~\cite{lohdefink2022adaptive}. Its key advantage lies in enabling end-to-end optimization: with additive noise or a Straight-Through Estimator (STE), gradients from both rate and distortion objectives can propagate back to the encoder. However, this element-wise design ignores inter-channel dependencies~\cite{gersho2012vector}, potentially limiting structural expressiveness.

In contrast, Vector Quantization (VQ)~\cite{vqgan,llamagen} maps groups of features to discrete indices via a learnable codebook, better preserving local structure and yielding more compact representations. This makes VQ particularly suitable for autoregressive modeling and extremely low-bitrate compression. However, the non-differentiable nearest-neighbor assignment blocks the gradient flow from the rate objective to the encoder, making joint RD optimization difficult. Consequently, most VQ-based methods~\cite{zhu2022unified,vqkmeans,uigc,lu2024hybridflow,xue2025one,wang2025switchable} focus on reconstruction quality while relying on indirect mechanisms for rate control.

\begin{figure*}[htbp]
  \centering
  \includegraphics[width=0.8\textwidth]{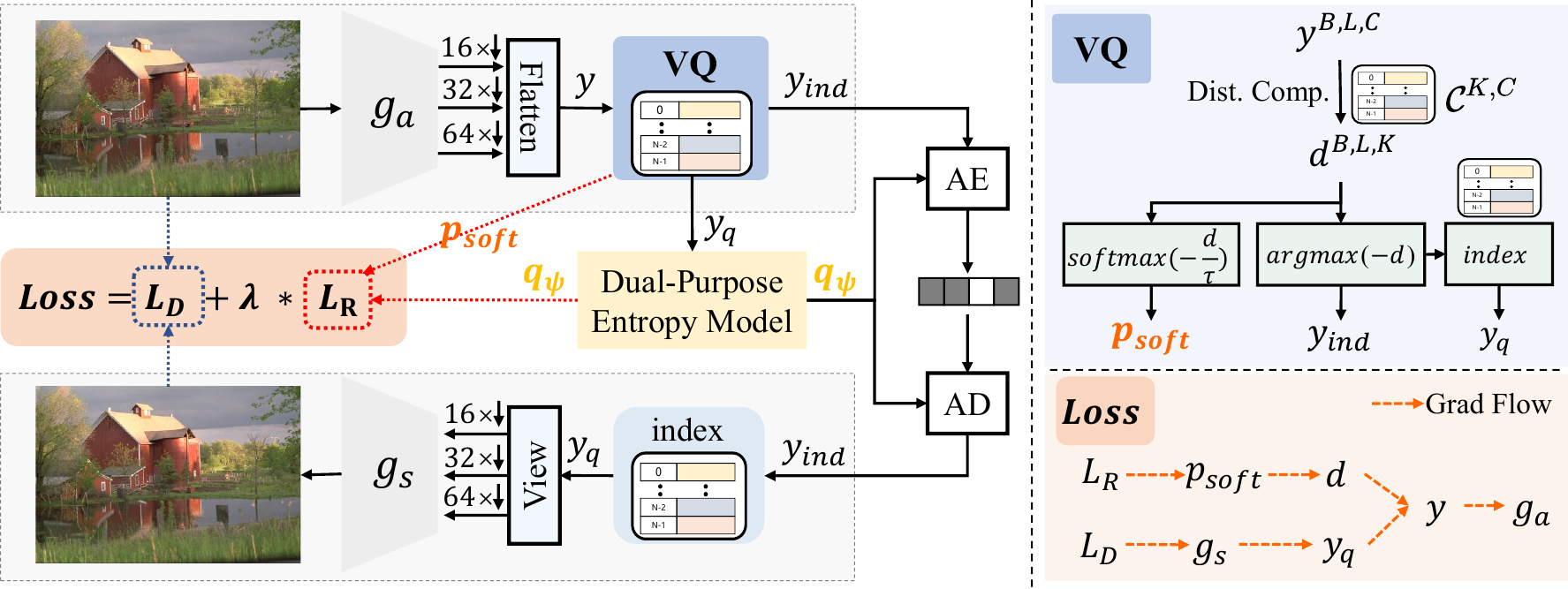} % 或设为0.9\columnwidth 等
  \vspace{-0.2cm}
  \caption{\textbf{Overview of RDVQ.} The analysis transform $\bm{g}_a$ extracts multi-scale features, which are flattened into a sequence $\bm{y}$ for vector quantization and entropy modeling. The VQ module produces hard-quantized embeddings $\bm{y}_q$, discrete indices $\bm{y}_{ind}$, and a relaxed distribution $p_{\text{soft}}$. During training, reconstruction is performed from $\bm{y}_q$, while the rate objective is computed as the cross-entropy between the relaxed distribution $p_{\text{soft}}$ and the entropy model prediction $q_{\bm{\psi}}$, enabling gradient flow from the rate loss to the encoder. The same model is used for entropy coding and supports test-time index completion. \textbf{Right:} Details of the VQ module and gradient flow. Hard assignment is used for reconstruction and coding, while the relaxed distribution is used only for rate estimation, restoring gradient flow from the rate loss ($L_R$) to the encoder $\bm{g}_a$.}

  % \vspace{-0.8cm}
  \label{fig:overview}
\end{figure*}

To mitigate this issue, existing approaches fall into two categories. Dual-branch methods~\cite{lu2024hybridflow,xue2025dlf,xue2025one,park2025diffo} combine VQ with SQ to incorporate entropy-based rate modeling, while single-branch methods~\cite{zhu2022unified,vqkmeans,uigc,ddcm,guo2025oscar} rely on VQ alone and adjust bitrate through heuristics such as codebook size or masking. Despite these efforts, both paradigms still fall short of enabling true end-to-end RD optimization in VQ-based compression.

\subsection{Generative image compression}
Generative image compression (GIC)~\cite{li2025onceforall,lee2024taco,mric,yang2023lossy,theis2022lossy,ma2024correcting,park2025diffo,agustsson2019generative,liang2025synonymous} leverages learned generative priors to reconstruct perceptually realistic images at extremely low bitrates~\cite{lei2023text+sketch,HiFiC,GLC,han2025generative,park2025diffo,relic2025bridging,korber2024egic}. In contrast to distortion-oriented codecs~\cite{VVC,liu2023img2,liu2025img3,jiang2023mlic++,mentzer2023m2t,QARV,li2025hpcm}, which often produce overly smooth results at low bitrates, GIC focuses on perceptual fidelity by restoring plausible textures.

Early works~\cite{mric,agustsson2019generative,HiFiC} introduced Generative Adversarial Networks (GANs)~\cite{GAN} into compression frameworks, significantly improving perceptual quality. More recently, diffusion-based models~\cite{ddcm_gen,ho2020denoising} have been incorporated into GIC, either in the image domain for texture refinement or in latent space for detail recovery, achieving strong perceptual performance~\cite{Perco,theis2022lossy,HiFiSD,korber2024egic,ghouse2023residual,xue2025one,guo2025oscar,diffeic, relic2025bridging}.

In addition to GANs and diffusion models, autoregressive (AR) models have demonstrated strong generative capabilities in image synthesis~\cite{vqgan,llamagen,zhang2026mvar} and restoration~\cite{varsr,li2025texture}. By modeling conditional distributions over discrete tokens, AR models are closely related to entropy modeling in compression. However, their potential for image compression, particularly in conjunction with structured representations such as VQ, remains less explored. This connection also suggests a natural mechanism for index completion under partial transmission. 
\section{Methodology}

\subsection{Problem Formulation: Joint RD Optimization}
\label{sec:prelim-lbic}

Modern learned image compression typically follows a variational autoencoder framework~\cite{balle2018variational}, where an analysis transform $\bm{g}_a$ maps an input image $\bm{x}$ to a latent representation $\bm{y}=\bm{g}_a(\bm{x})$, and a synthesis transform $\bm{g}_s$ reconstructs the image from its quantized version $\hat{\bm{y}}$.

The training objective is to minimize the rate--distortion (RD) Lagrangian
\begin{equation}
  \mathcal{L} = \lambda\, R + D(\bm{x}, \hat{\bm{x}}),
  \label{eq:rd_loss}
\end{equation}
where \(D(\cdot)\) measures reconstruction distortion and \(R\) denotes the expected coding rate, which is defined as
\begin{equation}
  R = \mathbb{E}_{\hat{\bm{y}}} \left[ -\log_{2} q_{\bm{\psi}}(\hat{\bm{y}}) \right],
\end{equation}
where \(q_{\bm{\psi}}\) is modeled by a learned entropy model.

Effective joint RD optimization requires gradients from both $D$ and $R$ to propagate back to the encoder. In scalar quantization (SQ)-based codecs, this is commonly achieved using differentiable approximations such as additive noise or the straight-through estimator, which provide an approximate gradient path through quantization.

In contrast, vector quantization (VQ) maps $\bm{y}$ to discrete codebook indices via nearest-neighbor assignment. Since the rate term is defined over these discrete indices, the mapping from $\bm{y}$ to $R$ is non-differentiable, and the rate objective cannot directly update the encoder. Therefore, the central challenge in VQ-based compression is to restore a differentiable rate-to-encoder pathway without changing the hard quantization process used in reconstruction and entropy coding.

\subsection{Overview of Proposed RDVQ} 
\label{Sec:framework_overview}

To overcome the missing rate-to-encoder gradient in VQ-based compression, we propose \textbf{RDVQ}, a unified framework that decouples the reconstruction/coding path from the rate-optimization path during training. RDVQ retains standard hard vector quantization for reconstruction and entropy coding, while introducing a differentiable soft relaxation in the rate-estimation branch.

As shown in Fig.~\ref{fig:overview}, the analysis transform $\bm{g}_a$ first extracts multi-scale latent features from the input image, which are then flattened into a unified representation $\bm{y}=\bm{g}_a(\bm{x})$ for vector quantization and entropy modeling. The VQ module produces
\begin{equation}
  \bm{y}_q, \bm{y}_{ind}, p_{\text{soft}} = \mathrm{VQ}(\bm{y}, \mathcal{C}),
\end{equation}
where $\mathcal{C}$ denotes the codebook, $\bm{y}_q$ the hard-quantized embeddings for reconstruction, $\bm{y}_{ind}$ the discrete indices for entropy coding, and $p_{\text{soft}}$ the relaxed distribution used only for rate optimization during training. The decoder reconstructs the image as $\hat{\bm{x}}=\bm{g}_s(\bm{y}_q)$, where $\bm{y}_q$ is reshaped back to its multi-scale feature structure.

To model the conditional distribution over codebook indices, RDVQ employs a \textbf{masked-Transformer entropy model}, which predicts conditional probabilities $q_{\bm{\psi}}$ by capturing both intra-scale spatial dependencies and inter-scale hierarchical dependencies. These probabilities are used not only for rate estimation and entropy coding, but also for generative index completion from partial observations.

During training, reconstruction is optimized using hard quantized features, while rate is optimized through the relaxed branch. At inference, the relaxation is removed and standard hard VQ is used for index selection and entropy coding; the same entropy model also enables limited-range test-time rate control via prefix transmission and autoregressive completion. Sec.~\ref{sec:diff_marginal} details the differentiable relaxation, and Sec.~\ref{sec:entropy_model} presents the entropy model.

\subsection{Differentiable Soft Relaxation}
\label{sec:diff_marginal}

To restore a differentiable rate-to-encoder pathway in RDVQ, we replace the hard index assignment only in the rate-estimation branch, while keeping hard vector quantization unchanged for reconstruction and entropy coding.

\begin{figure}[t]     
  \centering
  \includegraphics[width=0.48\textwidth]{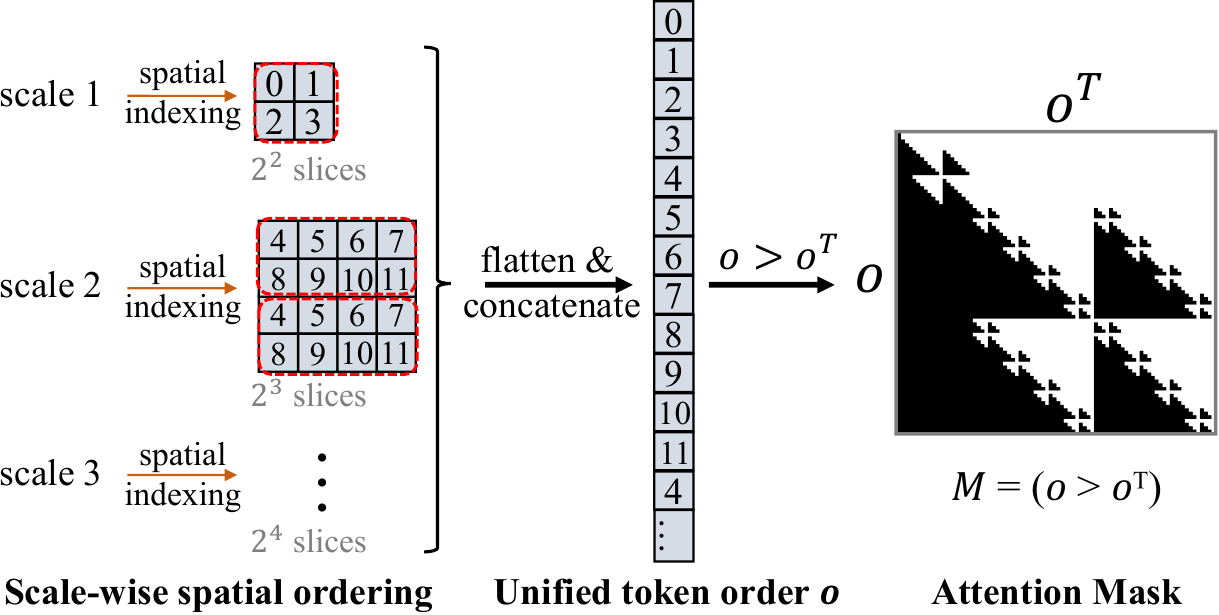} % 或设为0.9\columnwidth 等
  \vspace{-0.6cm}
  \caption{\textbf{Dependency-aware ordering for autoregressive entropy modeling.}
Scale-wise spatial orders are defined within each scale and concatenated into a unified order vector $\bm{o}$, from which the attention mask $\bm{M}=(\bm{o}>\bm{o}^{\top})$ is constructed.}
  % \vspace{-0.8cm}
  \label{fig:ms_feature_indexing}
\end{figure}

\noindent\textbf{From Hard Assignment to Soft Distribution.}
In standard VQ, each latent vector $\bm{y}_{b,l}$ is assigned to its nearest codebook entry,
\begin{equation}
  \bm{y}_{ind}(b,l) = \arg\min_{k} \| \bm{y}_{b,l} - \mathcal{C}_k \|^2,
\end{equation}
which is discrete and non-differentiable. We instead construct a continuous relaxation over codebook entries. Given encoder output $\bm{y} \in \mathbb{R}^{B\times L\times C}$ and codebook $\mathcal{C} \in \mathbb{R}^{K\times C}$, we first compute
\begin{equation}
  d_{b,l,k} = \| \bm{y}_{b,l} - \mathcal{C}_k \|^2,
\end{equation}
and then define a distance-aware soft distribution
\begin{equation}
  p_{\text{soft}}(b,l,k) = \operatorname{softmax}_k \left( -\frac{d_{b,l,k}}{\tau} \right),
  \label{eq:soft_posterior}
\end{equation}
where $\tau$ controls the sharpness of the relaxation. As $\tau \rightarrow 0$, $p_{\text{soft}}$ approaches the one-hot hard assignment while remaining differentiable for finite $\tau$.

\noindent\textbf{Relaxed Rate Objective.}
Using $p_{\text{soft}}$, we define the training-time surrogate rate
\begin{equation}
  R_{\text{soft}} = \mathbb{E}_{b,l} \left[ - \sum_{k=1}^{K} p_{\text{soft}}(b,l,k)\, \log q_{\bm{\psi}}(b,l,k) \right],
  \label{eq:rate_ce}
\end{equation}
where $q_{\bm{\psi}}$ is the entropy model over codebook indices. This cross-entropy serves as a differentiable proxy for the true coding cost.

\noindent\textbf{Gradient Bridge.}
The key property of this formulation is that
\[
\frac{\partial R_{\text{soft}}}{\partial \bm{y}} \neq 0,
\]
so the rate objective can directly shape the encoder representation. In contrast to standard VQ, where the rate term is defined on discrete indices and cannot provide usable gradients to $\bm{y}$, the relaxed objective encourages encoder outputs that are easier to predict under the entropy model and thus more compressible.

\noindent\textbf{Training--Inference Consistency.}
The relaxation is used only for rate estimation during training. Reconstruction always relies on hard quantized embeddings $\bm{y}_q$, and inference uses standard hard VQ for both index selection and entropy coding. Therefore, the proposed relaxation acts purely as a training surrogate, without changing the deployment pipeline of conventional VQ codecs. Multi-scale features are indexed to capture intra-scale and inter-scale conditional dependencies, and then flattened into a single concatenated sequence $\bm{o}$, which is used to generate attention mask and conditional input for the entropy model.

\begin{figure*}[htbp]     
  \centering
  \includegraphics[width=0.98\textwidth]{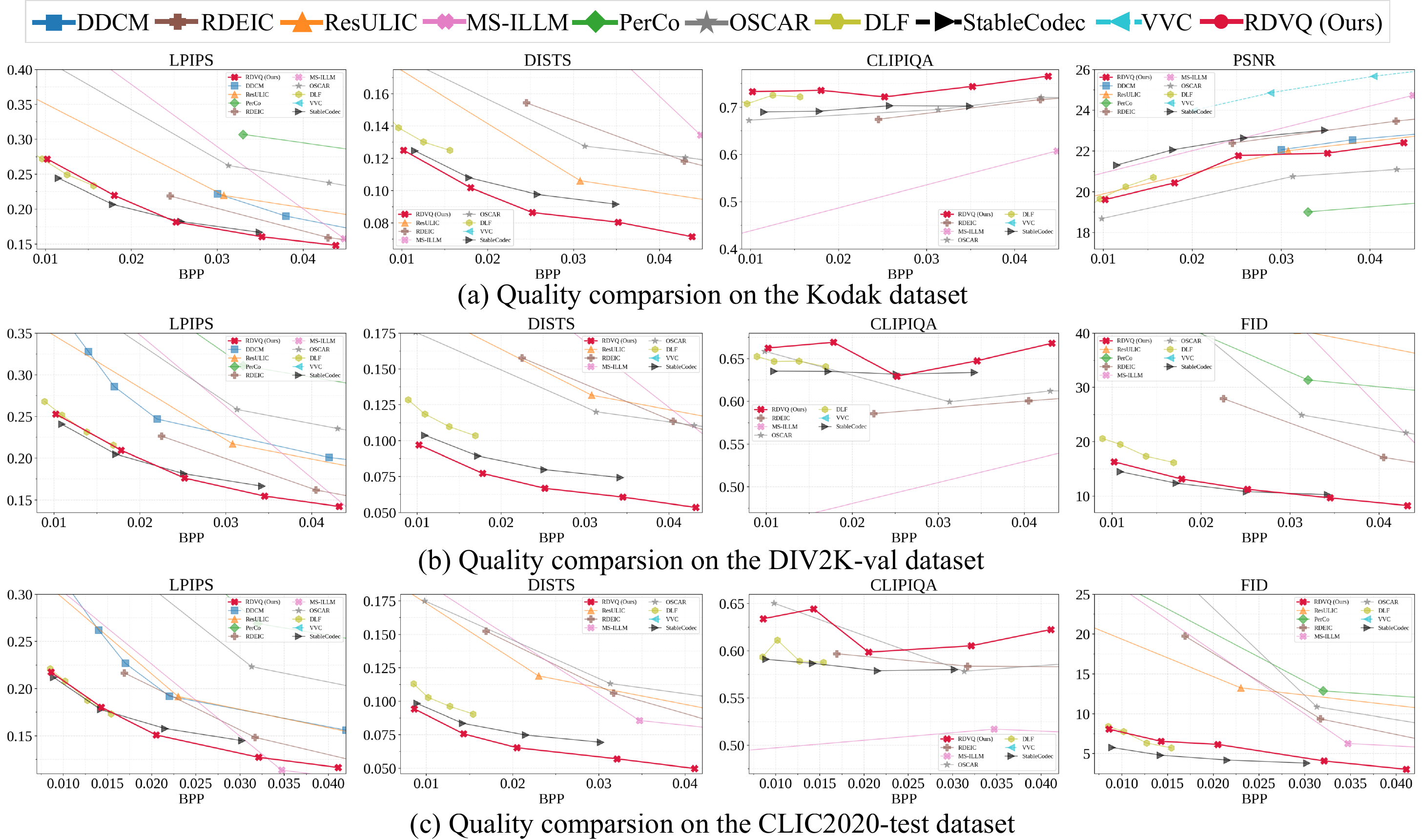} % 或设为0.9\columnwidth 等
  \vspace{-0.2cm}
  \caption{Rate-distortion curves on the Kodak, the DIV2k-val and the CLIC2020-test datasets.}
  % \vspace{-0.8cm}
  \label{Fid.RD_curve}
\end{figure*}

\subsection{Autoregressive Entropy Prediction}
\label{sec:entropy_model}

To optimize the relaxed rate objective in Eq.~\eqref{eq:rate_ce}, RDVQ relies on an accurate conditional model of codebook indices. We therefore employ a unified \textbf{masked-Transformer entropy model} to predict the conditional probabilities $q_{\bm{\psi}}$ over codebook indices by modeling structured dependencies among multi-scale tokens.

\noindent\textbf{Dependency-Aware Token Ordering.}
The encoder features are inherently multi-scale and contain both intra-scale spatial dependencies and inter-scale hierarchical dependencies. Although they are flattened into a unified token sequence for quantization and entropy modeling, this underlying structure still defines the desired causal relationships among tokens. We therefore exploit the original scale-wise organization to construct a dependency-aware order over all index tokens. As shown in Fig.~\ref{fig:ms_feature_indexing}, tokens are grouped and ordered spatially within each scale, with finer scales partitioned more finely to reflect their richer local structure, and are arranged across scales in a coarse-to-fine manner so that finer-scale tokens are conditioned on preceding coarser-scale ones. The resulting per-scale orders are flattened and concatenated into a unified order vector $\bm{o}$.

\noindent\textbf{Masked Autoregressive Modeling.}
Based on $\bm{o}$, we construct the attention mask
\begin{equation}
    \bm{M} = (\bm{o} > \bm{o}^{\top}),
\end{equation}
which allows each token to attend only to its valid predecessors. The quantized tokens are reordered according to $\bm{o}$ so that the input sequence is aligned with the predefined causal order. Together, the reordered input and the mask $\bm{M}$ enable parallel training while preserving the intended autoregressive factorization over codebook indices.

\noindent\textbf{Unified Role in Entropy Coding and Rate Control.}
The predicted probabilities $q_{\bm{\psi}}$ play a dual role in RDVQ. They are used as the entropy model for rate estimation during training and for entropy coding at inference. In addition, given a transmitted prefix, the same model autoregressively completes the remaining suffix indices, enabling \emph{limited-range test-time rate control} without retraining by varying the prefix length. Compared with prior completion-based methods such as UIGC~\cite{uigc}, our predictor is directly coupled to the learned index distribution used in rate--distortion optimization, leading to better-calibrated entropy estimates within the supported operating range.
\section{Experiments}

\subsection{Experimental Settings}

\begin{figure*}[htbp]     
  \centering
  \includegraphics[width=0.92\textwidth]{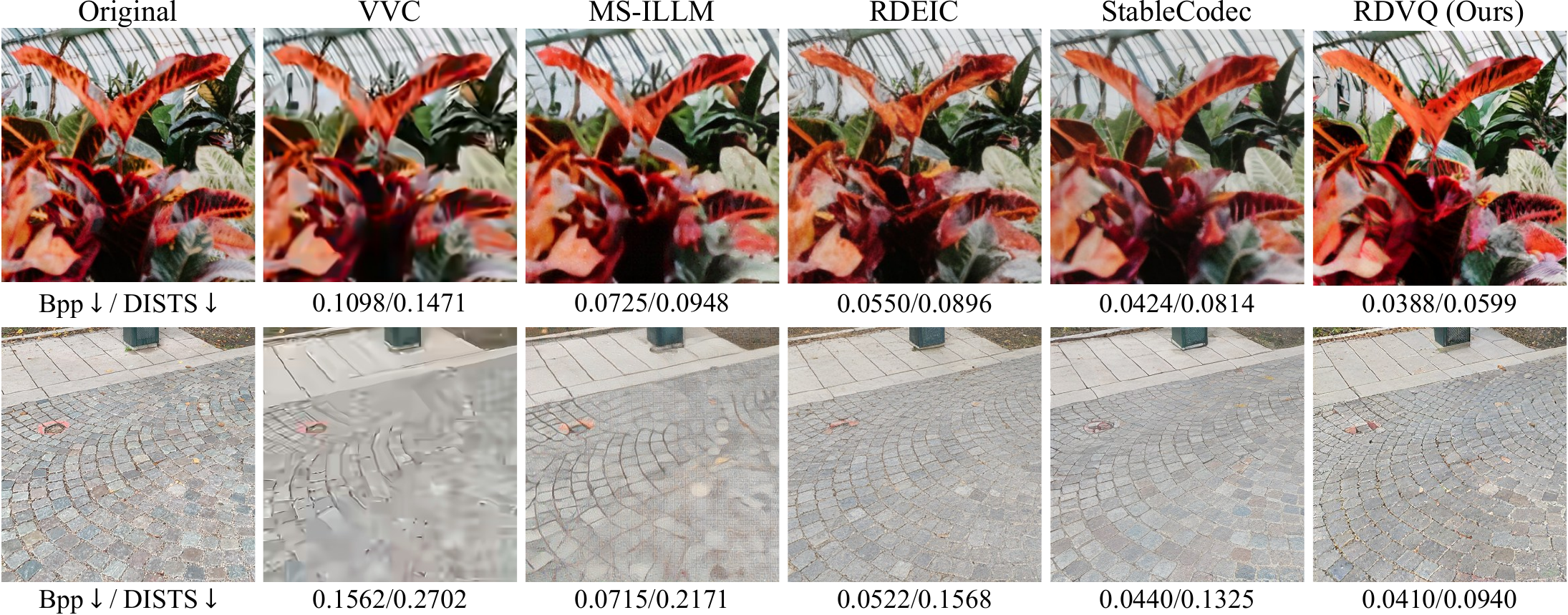} % 或设为0.9\columnwidth 等
  \vspace{-0.2cm}
  \caption{Visual examples on the CLIC2020 test set. Zoom in for better view.}
  \vspace{-0.2cm}
  \label{fig:CLIC2020 comparsion}
\end{figure*}

\noindent \textbf{Model and Training.} 
RDVQ comprises an autoencoder and an autoregressive entropy model. The autoencoder is adapted from the VQ-VAE architecture in LlamaGen~\cite{llamagen}, where attention layers are removed and the feature decomposition is extended to a multi-scale design to capture richer structural information. The entropy model is implemented using standard transformer layers with the proposed masking mechanism, enabling autoregressive prediction of codebook indices and accurate rate estimation.

We optimize both reconstruction quality and coding rate using a three-stage training schedule: (i) pretraining the autoencoder and codebook on ImageNet with reconstruction losses; (ii) pretraining the entropy model with the rate objective; and (iii) joint fine-tuning of the full model with both objectives, followed by high-resolution adaptation on OpenImage~\cite{openimage} and DF2K~\cite{Div2k} to support multi-resolution inference. Detailed architectures and training hyperparameters are provided in the supplementary material.

\begin{figure}[t]     
  \centering
  \includegraphics[width=0.38\textwidth]{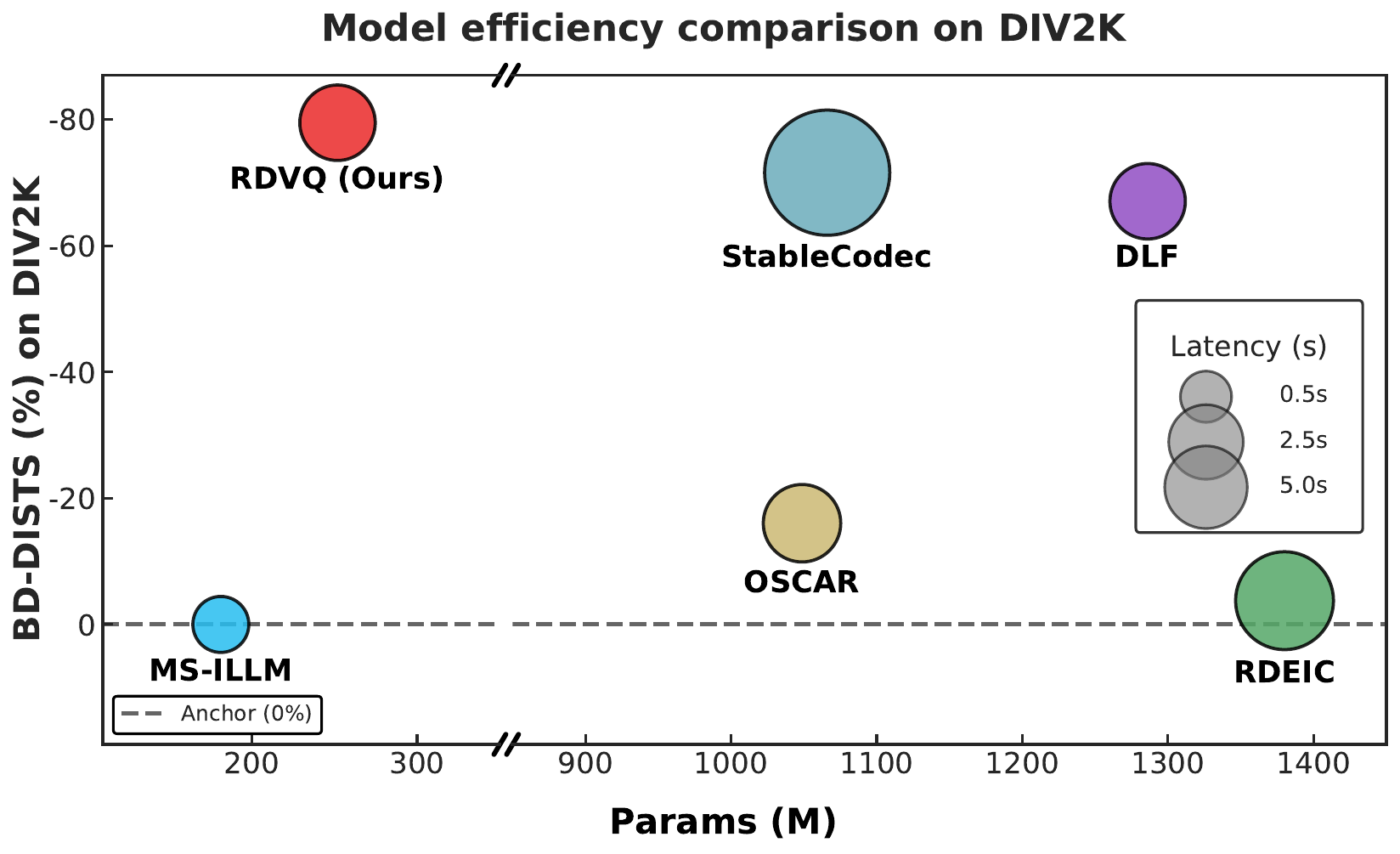} % 或设为0.9\columnwidth 等
  \vspace{-0.3cm}
  \caption{Model efficiency comparison on DIV2K~\cite{Div2k}. RDVQ achieves the best BD-DISTS with less than 20\% of the parameters of most baselines, while maintaining competitive latency.}
  \label{fig:Efficiency}
  \vspace{-0.3cm}
\end{figure}

\noindent \textbf{Evaluation Datasets.}
Following prior work~\cite{guo2025oscar,resulic,ddcm}, we evaluate on three standard benchmarks: Kodak~\cite{Kodak}, CLIC2020 Test~\cite{CLIC2020}, and the DIV2K validation set~\cite{Div2k}. Kodak contains 24 natural images at $768 \times 512$ resolution. CLIC2020 Test and DIV2K validation include 428 and 100 high-quality 2K images, respectively, covering diverse scenes and textures. All results are obtained using full-resolution inference.

\noindent \textbf{Evaluation Metrics.}
We assess perceptual quality using both supervised and unsupervised metrics. Supervised metrics include DISTS~\cite{DISTS}, LPIPS~\cite{lpips}, and FID~\cite{FID}, evaluating structural fidelity, perceptual similarity, and distributional realism, respectively (FID is omitted on Kodak due to its limited size). Unsupervised metric CLIPIQA~\cite{CLIPIQA} measures perceptual quality and semantic consistency without reference images. Bitrate is reported in bits per pixel (bpp).

\begin{figure*}[htbp]     
  \centering
  \includegraphics[width=0.92\textwidth]{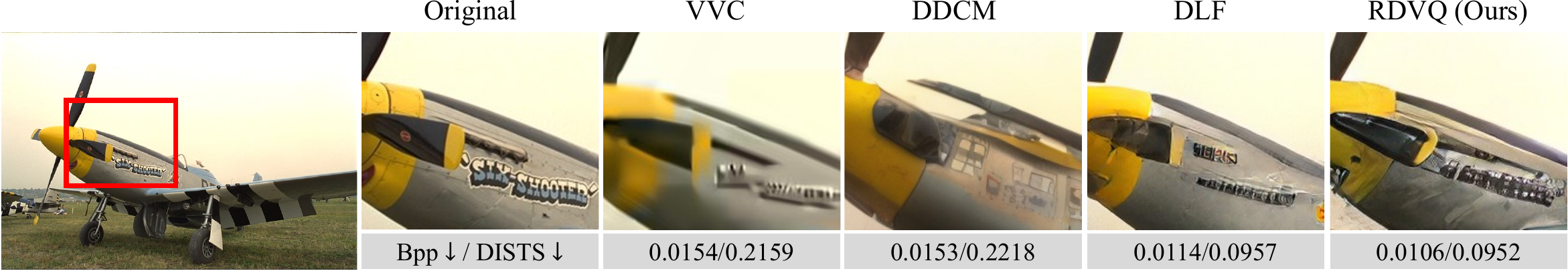} % 或设为0.9\columnwidth 等
  \vspace{-0.2cm}
  \caption{Visual examples on the Kodak dataset.}
  % \vspace{-0.8cm}
  \label{fig:kodak comparsion}
  \vspace{-0.3cm}
\end{figure*}

\noindent \textbf{Comparison Methods.}
We compare RDVQ with a traditional codec and recent GIC methods, grouped by quantization strategy. As a conventional baseline, we include VVC~\cite{VVC}. GIC methods are categorized into scalar quantization (SQ): MS-ILLM~\cite{msillm}, PerCO~\cite{Perco}, ResULIC~\cite{resulic}, StableCodec~\cite{zhang2025stablecodec}; vector quantization (VQ): DDCM~\cite{ddcm}, OSCAR~\cite{guo2025oscar}; and hybrid SQ-VQ: DLF~\cite{xue2025dlf}, RDEIC~\cite{li2025rdeic}. This grouping highlights the role of quantization in rate control and reconstruction behavior. Notably, unlike most baselines (except MS-ILLM) that rely on large pretrained models (\eg, diffusion~\cite{ddcm} or ViT~\cite{vit}), RDVQ is trained from scratch using only GAN~\cite{GAN} and perceptual losses~\cite{lpips}.

\subsection{Main Results}

\noindent \textbf{Quantitative Analysis.}
Fig.~\ref{Fid.RD_curve} presents rate--distortion (R--D) curves on Kodak~\cite{Kodak}, DIV2K-val~\cite{Div2k}, and CLIC2020-test~\cite{CLIC2020} under ultra-low bitrate settings. Despite being trained from scratch with only GAN~\cite{GAN} and perceptual losses~\cite{lpips}, RDVQ achieves state-of-the-art performance on DISTS~\cite{DISTS} and CLIPIQA~\cite{CLIPIQA} across all datasets. In addition, RDVQ delivers LPIPS and FID~\cite{FID} results that are highly competitive with, and often superior to, methods built upon large-scale pretrained priors (\eg, OSCAR~\cite{guo2025oscar}, RDEIC~\cite{li2025rdeic}, and ResULIC~\cite{resulic}). These consistent gains across both supervised and unsupervised metrics demonstrate the effectiveness of our joint rate--distortion optimization in preserving structural fidelity and perceptual realism.

Beyond compression performance, RDVQ also exhibits strong efficiency advantages. As shown in Fig.~\ref{fig:Efficiency}, which plots BD-DISTS against model size on DIV2K-val, RDVQ achieves the best BD-DISTS while using less than 20\% of the parameters of most baselines. Compared with foundation-model-based methods such as StableCodec~\cite{zhang2025stablecodec} and DLF~\cite{xue2025dlf}, RDVQ attains superior performance with significantly reduced model complexity (251.9M parameters). Moreover, RDVQ maintains competitive inference speed, processing a 2K image in 1.3 seconds on an NVIDIA RTX 4090 GPU. These results show that RDVQ is both lightweight and efficient.

\begin{figure}[t]     
  \centering
  \includegraphics[width=0.35\textwidth]{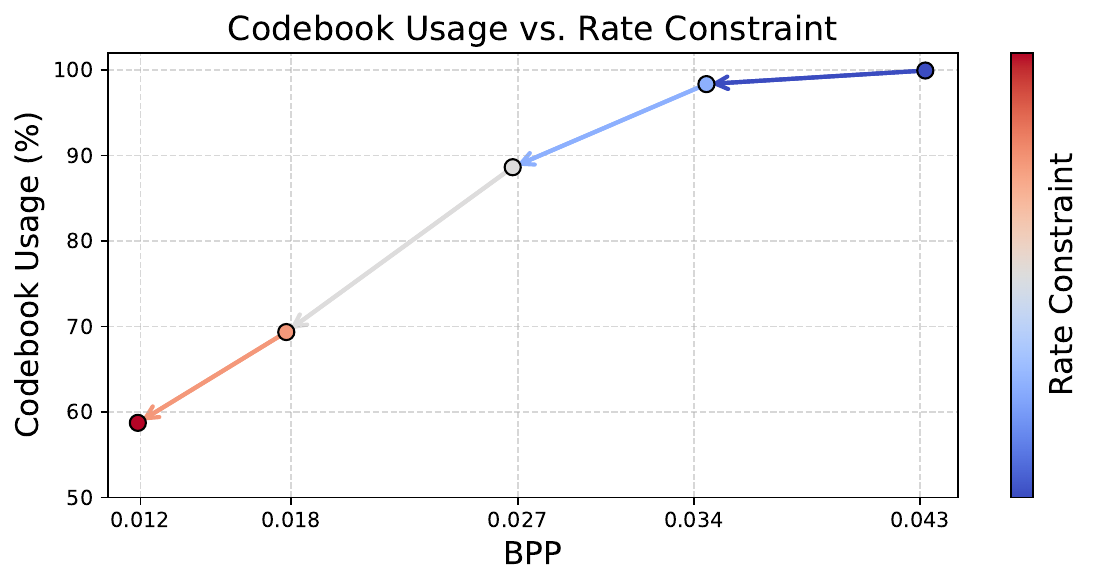} % 或设为0.9\columnwidth 等
  \vspace{-0.3cm}
  \caption{Codebook usage becomes more concentrated under stronger rate constraints.}
  \label{fig:CodebookUsage}
  \vspace{-0.4cm}
\end{figure}

\begin{figure*}[htbp]     
  \centering
  \includegraphics[width=0.85\textwidth]{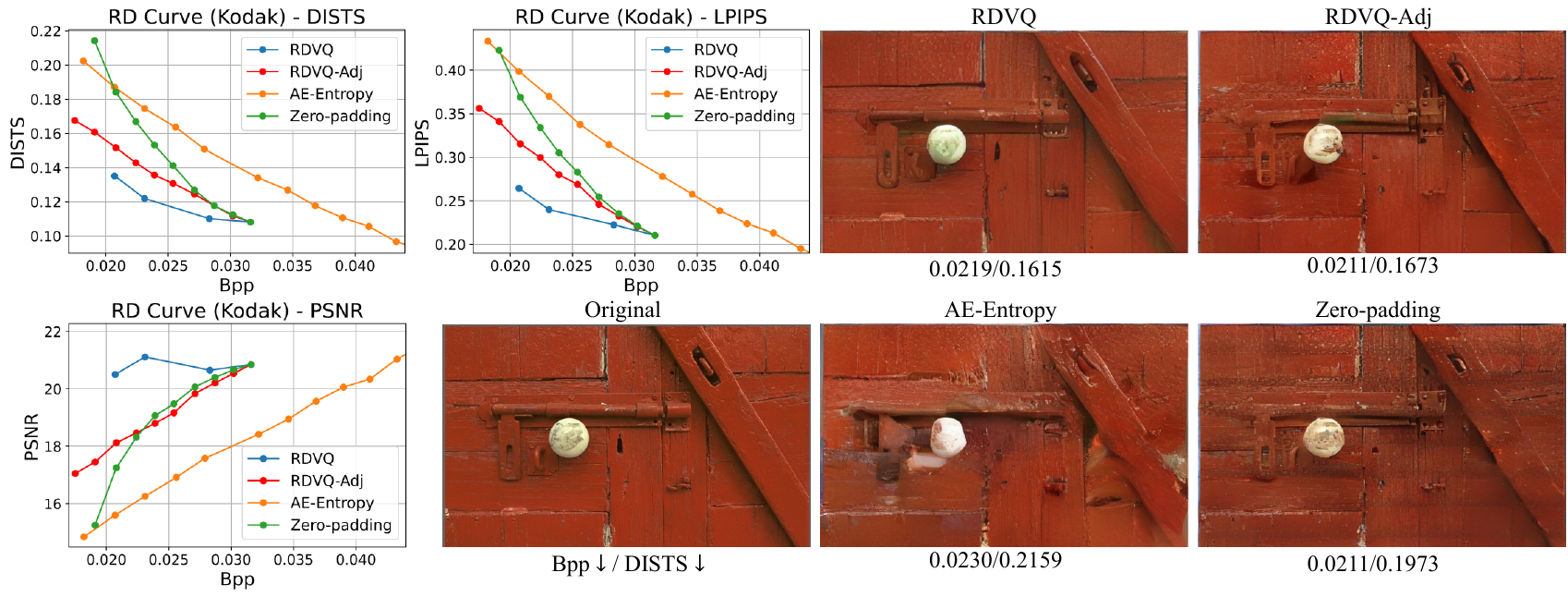} % 或设为0.9\columnwidth 等
  \vspace{-0.3cm}
  \caption{Test-time rate adjustment via prefix transmission. \textbf{Left:} Rate--distortion curves on Kodak. \textbf{Right:} Visual comparisons at similar bitrates (Bpp / DISTS). 
RDVQ-Adj maintains competitive performance with gradual degradation, while AE-Entropy and Zero-padding suffer from more noticeable artifacts.}
  % \vspace{-0.8cm}
  \label{fig:rate_adjust}
  \vspace{-0.35cm}
\end{figure*}

\noindent \textbf{Qualitative Evaluation.}
Visual comparisons on CLIC and Kodak are shown in Fig.~\ref{fig:CLIC2020 comparsion} and Fig.~\ref{fig:kodak comparsion}. At ultra-low bitrates (Kodak, down to 0.01 bpp), RDVQ preserves fine structural edges and repetitive textures with minimal artifacts, producing reconstructions that remain faithful to the reference. In contrast, VVC suffers from over-smoothing and detail loss, diffusion-based DDCM tends to generate hallucinated textures inconsistent with the ground truth, and the hybrid SQ--VQ method (DLF) exhibits incomplete detail recovery with weakened local structures.

This advantage persists at slightly higher bitrates (CLIC, 0.03$\sim$0.04 bpp), where RDVQ continues to better preserve both structural geometry and fine textures. Other learned codecs (\ie, MS-ILLM, RDEIC, OSCAR) show varying degrees of structural distortion or texture degradation, while VVC still produces blurred results. Additional visual comparisons are provided in the supplementary material.

\noindent \textbf{Test-Time Rate Adjustment.} 
RDVQ supports test-time rate adjustment within a limited range by transmitting only a prefix of the discrete index sequence, while autoregressively predicting the remaining indices using the entropy model. This design enables flexible bitrate control without retraining, providing a practical trade-off between compression efficiency and perceptual quality.

We conduct controlled comparisons under a shared lightweight backbone to isolate the effect of the proposed framework. Specifically, we evaluate: (a) \textbf{RDVQ}, using full index transmission; (b) \textbf{RDVQ-Adj}, enabling prefix-based transmission with autoregressive completion; (c) \textbf{AE-Entropy}, where the autoencoder and entropy model are optimized separately; and (d) \textbf{Zero-padding}, which replaces missing features with zeros.

As shown in Fig.~\ref{fig:rate_adjust}, RDVQ-Adj exhibits a smooth and gradual degradation along the R--D curve within the range of 0.02--0.32 bpp, closely tracking the baseline RDVQ. This behavior indicates that the jointly optimized latent space remains highly predictable under partial transmission, allowing stable completion while preserving perceptual quality. 

In contrast, AE-Entropy and Zero-padding suffer from significantly larger quality drops and visible artifacts, reflecting the lack of predictability in their latent representations. Both quantitative and qualitative results consistently show that RDVQ-Adj achieves a more favorable rate--distortion trade-off with cleaner structures and fewer artifacts. These findings highlight that joint rate--distortion optimization is essential for enabling effective test-time rate adaptation within a limited operating range.

\begin{table}[t]
\centering
\caption{Quantitative comparison of ablation variants on DIV2K-val. 
Lower DISTS, LPIPS, FID indicate better perceptual quality.}
\vspace{0.1em}
\resizebox{\linewidth}{!}{
\begin{tabular}{l|ccccc}
\toprule
\textbf{Method} & \textbf{bpp} $\downarrow$ & \textbf{DISTS} $\downarrow$ & \textbf{LPIPS} $\downarrow$ & \textbf{FID} $\downarrow$ \\
\midrule
\textbf{RDVQ (full)} & \textbf{0.0247} & \textbf{0.1005} & \textbf{0.2321} & \textbf{19.96} \\
\textbf{w/o Relaxation} & 0.0464 & 0.2147 & 0.5031 & 86.93\\
\textbf{K-means VQ} & 0.0247 & 0.1253 & 0.2831 & 28.08\\
\bottomrule
\end{tabular}
}
\label{tab:ablation}
\vspace{-0.2cm}
\end{table}

\subsection{Analysis and Ablation Studies}
\label{sec:ablation}

\noindent \textbf{Analysis of R--D Optimization.}
To analyze the effect of joint rate--distortion (R--D) optimization, we examine how encoder representations and codebook utilization evolve across different bitrate levels.

\begin{figure}[t]     
  \centering
  \includegraphics[width=0.49\textwidth]{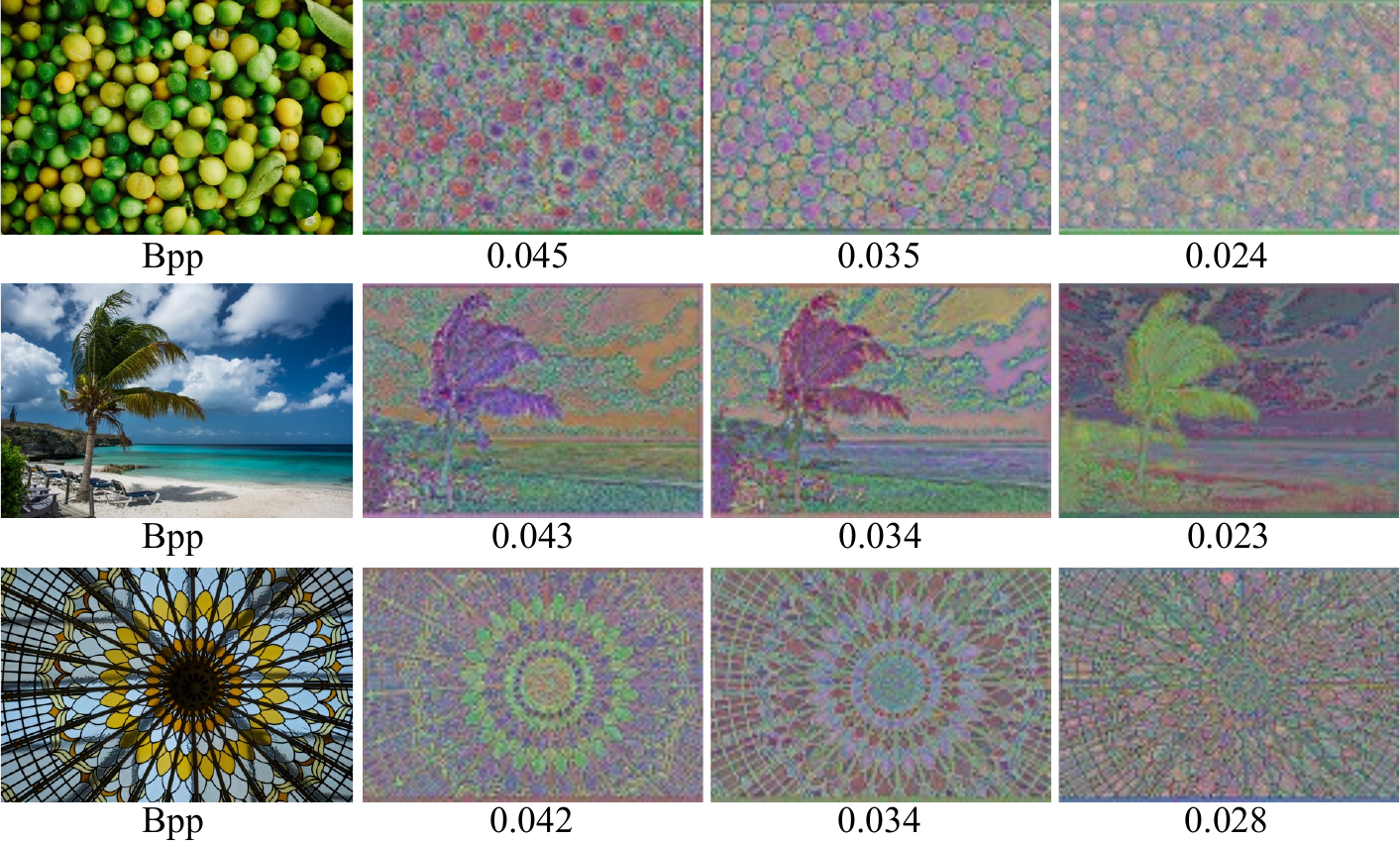} % 或设为0.9\columnwidth 等
  \vspace{-0.6cm}
  \caption{PCA visualization of the largest-scale encoder features under different compression ratios. As the bitrate decreases, the features become progressively smoother, with high-frequency details suppressed.}
  \label{fig:PCA_vsiual}
  \vspace{-0.35cm}
\end{figure}

We first visualize the encoder features using PCA on the largest-scale representations, mapped to RGB space in Fig.~\ref{fig:PCA_vsiual}. As the bitrate decreases, the features progressively emphasize smooth, low-frequency structures while suppressing high-frequency details. This behavior indicates that the encoder adapts to compression constraints by focusing on more predictable and compact representations, enabling efficient coding while preserving perceptual quality.

We further analyze codebook utilization in Fig.~\ref{fig:CodebookUsage}. With decreasing bitrate, the utilization becomes increasingly concentrated on a smaller subset of codebook entries. This trend suggests that the model reduces redundancy by selectively activating the most representative atoms, improving coding efficiency under tighter rate constraints.

Overall, these observations demonstrate that joint R--D optimization encourages the model to learn more predictable feature representations and more efficient codebook usage, which together contribute to improved compression performance at low bitrates.

\noindent \textbf{Ablation Studies.}
We evaluate the effectiveness of the proposed differentiable relaxation and joint R--D optimization by comparing RDVQ with two representative alternatives. All variants share the same lightweight backbone and are trained under identical settings.

As shown in Table~\ref{tab:ablation}, removing the differentiable relaxation (\textbf{w/o Relaxation}) leads to a severe performance drop, even at a substantially higher bitrate. In this case, the rate loss can only propagate to the encoder indirectly through the entropy model prediction, resulting in a weak and unstable optimization signal. This confirms that a differentiable index distribution is essential for enabling effective end-to-end R--D optimization.

We further compare with a K-means-based rate control strategy (\textbf{K-means VQ}~\cite{vqkmeans}), which adjusts bitrate by clustering the codebook and varying its size. While this approach achieves the same bitrate as RDVQ, it yields noticeably worse perceptual quality across all metrics. This suggests that heuristic rate control via codebook size does not eliminate redundancy in the index distribution, leading to suboptimal compression efficiency.

Overall, these results demonstrate that the proposed relaxation enables stable rate optimization, while joint R--D learning effectively improves representation efficiency beyond heuristic VQ-based methods.

\section{Conclusion}
We present RDVQ, a differentiable rate–distortion optimization framework for VQ-based image compression. By introducing a soft index distribution for rate estimation, our method enables effective end-to-end R–D optimization and more efficient codebook utilization. Extensive experiments demonstrate that RDVQ achieves superior perceptual quality at extremely low bitrates with a lightweight architecture. Moreover, our formulation provides a principled way to bridge discrete representation learning and continuous optimization, offering improved stability and flexibility for rate control. These results highlight the importance of explicit rate modeling in advancing generative image compression and suggest its potential for broader applications in discrete generative modeling.

\section*{Acknowledgement}
This work was primarily supported by the National Natural Science Foundation of China (No.~62476051), and partially supported by the Key Project of the Natural Science Foundation of Sichuan Province (Grant 2025ZNSFSC0002).
{
    \small
    \bibliographystyle{ieeenat_fullname}
    \bibliography{main}
}

\clearpage

\setcounter{page}{1}
\setcounter{section}{0}
\renewcommand{\thesection}{\Alph{section}}
\maketitlesupplementary

\def\@seccntformat#1{\csname the#1\endcsname\quad}

\renewcommand{\thefigure}{S\arabic{figure}}
\renewcommand{\thetable}{S\arabic{table}}

\setcounter{figure}{0}
\setcounter{table}{0}

\section{Architecture Details}
\label{sec:appendix_architecture}

RDVQ consists of two main components: an \emph{image tokenizer} that converts images into discrete latent tokens for representation and reconstruction, and a \emph{masked-Transformer entropy model} that predicts their conditional distributions for rate estimation and entropy coding. We describe each component in detail below.

\paragraph{Image Tokenizer.}
The image tokenizer is a vector-quantized autoencoder adapted from the LlamaGen~\cite{llamagen} architecture, designed to map images into discrete codebook indices while enabling high-quality reconstruction. It consists of an encoder, a decoder, and a shared codebook. To better support compression-oriented rate--distortion (RD) optimization and cross-resolution generalization, we introduce two key modifications: (i) all attention layers are removed for improved efficiency and scalability, and (ii) the original single-scale latent representation is extended to a multi-scale hierarchy.

\emph{Codebook.}
A shared codebook of size $4096 \times 32$ is used across all scales. The codebook is learned during a first-stage pretraining phase and then fixed during RD optimization to stabilize training.

\emph{Multi-scale representation.}
Instead of producing a single latent map at a fixed downsampling ratio, the encoder extracts features at three scales with downsampling factors of $16$, $32$, and $64$. These multi-scale latents form a coarse-to-fine representation, where the lowest-resolution features capture global structure and higher-resolution features encode progressively finer details. To encourage a clean hierarchical decomposition, each higher-scale latent is defined as the residual between its feature map and the upsampled lower-scale feature. 

This design not only introduces structured dependencies across scales, which benefits entropy modeling, but also increases the upper bound of achievable bitrate. For example, for a $256\times256$ image, the maximum bitrate under uniform coding is given by
\begin{equation}
\begin{split}
\text{bpp}_{\max} 
= \frac{(4^2 + 8^2 + 16^2) \cdot \log_2(4096)}{256^2}
\approx 0.0615,
\end{split}
\end{equation}
which is higher than that of a single-scale tokenizer, enabling a wider RD operating range,.

\emph{Network architecture.}
Both the encoder and decoder consist of six stages of downsampling or upsampling. Each stage contains two ResBlocks followed by a downsampling or upsampling operator. The channel dimensions across stages are set to $128 \!\times\! [1, 1, 2, 2, 4, 4, 4]$. During encoding, latent features are extracted from the three lowest-resolution stages to construct the multi-scale representation, and the decoder reconstructs the image from the corresponding quantized embeddings.

\paragraph{Entropy Model.}
We employ a masked Transformer to predict a categorical distribution over codebook indices for each token, which is used for rate estimation during training and entropy coding at inference. The model operates on quantized latent tokens produced by the image tokenizer.

\emph{Architecture.}
The model consists of an input projection, a Transformer backbone, and an output projection. The input projection maps 32-dimensional quantized embeddings to 768-dimensional tokens. The backbone contains 12 masked Transformer layers with 8 attention heads and an MLP expansion ratio of 4. The output projection is a linear classifier over 4096 codebook entries.

\emph{Masking strategy.}
To model both intra-scale spatial dependencies and inter-scale hierarchical dependencies, we construct a dependency-aware autoregressive order over multi-scale latent features (coarse to fine).

As shown in Fig.~\ref{fig:ms_feature_indexing}, at each scale $i$, spatial positions are evenly partitioned into $n_i = 2^{(i+1)}$ segments, and each token is assigned a decoding number indicating its relative order within the scale. To enforce cross-scale dependencies, decoding numbers at finer scales are offset by the cumulative counts from all coarser scales (\eg, adding $n_1$ for scale 2 and $n_1+n_2$ for scale 3), ensuring that finer-scale tokens are conditioned on all coarser-scale tokens.

The decoding numbers are then flattened into a global order vector $o$, from which the attention mask is defined as
\begin{equation}
    M = o > o^T,
\end{equation}
so that each token attends only to its valid predecessors.

Tokens are reordered according to $o$ before being fed into the Transformer. Multi-scale quantized features are rearranged to follow this order, where each token is conditioned on its preceding context: the first slice at scale $i$ is initialized from the previous scale (or a learnable token for $i=1$), and subsequent slices use the preceding feature as input. This ensures consistency between the input sequence and the causal mask, enabling autoregressive modeling of both intra- and inter-scale dependencies.

\emph{Resolution generalization.}
To support variable input resolutions, we first partition the latent feature map at each scale into non-overlapping spatial windows (\ie, $4\times4$, $8\times8$, and $16\times16$ for the three scales). For each spatial location, windows from different scales that correspond to the same image region are then grouped together, and their tokens are concatenated into a single sequence for processing. Groups from different spatial locations are treated independently and batched for parallel computation, enabling flexible resolution handling while preserving aligned local multi-scale structure.

\begin{figure*}[t]     
  \centering
  \includegraphics[width=0.77\textwidth]{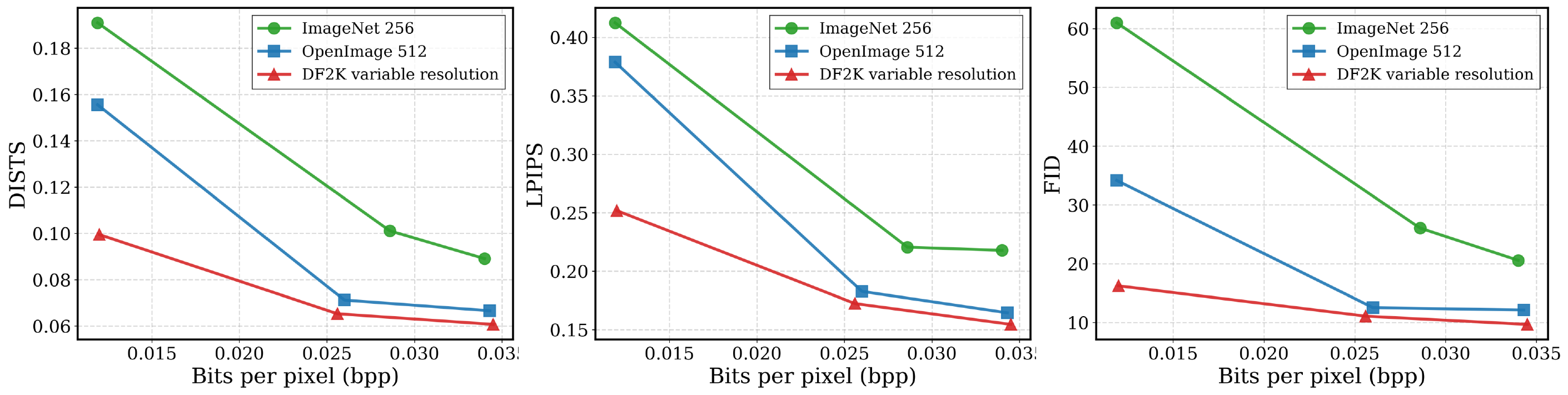} % 或设为0.9\columnwidth 等
  \vspace{-0.3cm}
  \caption{Effectiveness of tuning on high-quality images (evaluated on Div2K validation set).}
  \label{fig:High_reso_adaption}
  \vspace{-0.25cm}
\end{figure*}

\section{Training Details and Hyperparameters}
\label{sec:training_details_sup}

\paragraph{Optimization objective.}
RDVQ is trained under a rate--distortion (RD) objective that jointly optimizes compression efficiency and reconstruction quality:
\begin{equation}
\mathcal{L} = \mathcal{L}_D + \lambda \mathcal{L}_R,
\end{equation}
where $\mathcal{L}_R$ denotes the rate loss and $\mathcal{L}_D$ denotes the distortion loss.

The rate loss is defined as a cross-entropy objective between the relaxed index distribution $p_{\text{soft}}$ and the entropy model prediction $q_{\bm{\psi}}$:
\begin{equation}
\mathcal{L}_R = \mathrm{CE}(p_{\text{soft}}, q_{\bm{\psi}}).
\end{equation}

The distortion loss is composed of multiple terms to balance fidelity and perceptual quality:
\begin{equation}
\mathcal{L}_D = \mathcal{L}_{\text{codebook}} + \mathcal{L}_{\text{MSE}} + \mathcal{L}_{\text{LPIPS}} + 0.1\mathcal{L}_{\text{GAN}}.
\end{equation}
Here, $\mathcal{L}_{\text{codebook}}$ denotes the vector quantization loss~\cite{vqgan}, $\mathcal{L}_{\text{MSE}}$ enforces pixel-wise reconstruction accuracy, $\mathcal{L}_{\text{LPIPS}}$~\cite{lpips} improves perceptual similarity, and $\mathcal{L}_{\text{GAN}}$~\cite{GAN} further enhances visual realism.

\paragraph{Training pipeline.}
To obtain a high-quality codebook and stabilize rate--distortion (RD) optimization, we adopt a three-stage training pipeline, where different objectives are applied progressively.

In the first stage, we pretrain the image tokenizer using only the distortion objective $\mathcal{L}_D$. This stage focuses on learning a representative codebook and high-quality reconstruction without considering compression efficiency.

In the second stage, we train the entropy model separately using only the rate objective $\mathcal{L}_R$. With the tokenizer and codebook fixed, this stage learns to accurately model the distribution over codebook indices, providing reliable probability estimates for subsequent RD optimization.

In the final stage, we perform joint RD optimization using the full objective $\mathcal{L} = \mathcal{L}_D + \lambda \mathcal{L}_R$, while keeping the codebook fixed. Models at different bitrate levels are obtained through a staged $\lambda$-based curriculum, which enables smooth adaptation across rate regimes.

After training on ImageNet, all models are further fine-tuned on higher-resolution datasets to improve perceptual quality and generalization.

\paragraph{Implementation details.}
We train RDVQ on ImageNet~\cite{imagenet}, OpenImage~\cite{openimage}, and DF2K~\cite{Div2k} datasets. ImageNet is used for large-scale low-resolution pretraining, while OpenImage and DF2K are employed for high-resolution fine-tuning to improve cross-resolution generalization during inference.

\textbf{Training stages.} The image tokenizer is first pretrained on $256\times256$ ImageNet patches for $7\!\times\!10^5$ iterations (batch size 32, learning rate $1\!\times\!10^{-4}$, BF16). The learned codebook is then frozen. Next, the entropy model is pretrained on $256\times256$ ImageNet patches for $4\!\times\!10^5$ iterations (batch size 80) to provide stable probability estimation. Finally, joint RD optimization is performed with the full objective $\mathcal{L} = \mathcal{L}_D + \lambda \mathcal{L}_R$ and the codebook fixed. Models are first trained at relatively high bitrates and then progressively fine-tuned toward lower bitrates. From this stage onward, all models are trained in FP32 precision for stable optimization.

\begin{figure*}[t]     
  \centering
  \includegraphics[width=0.77\textwidth]{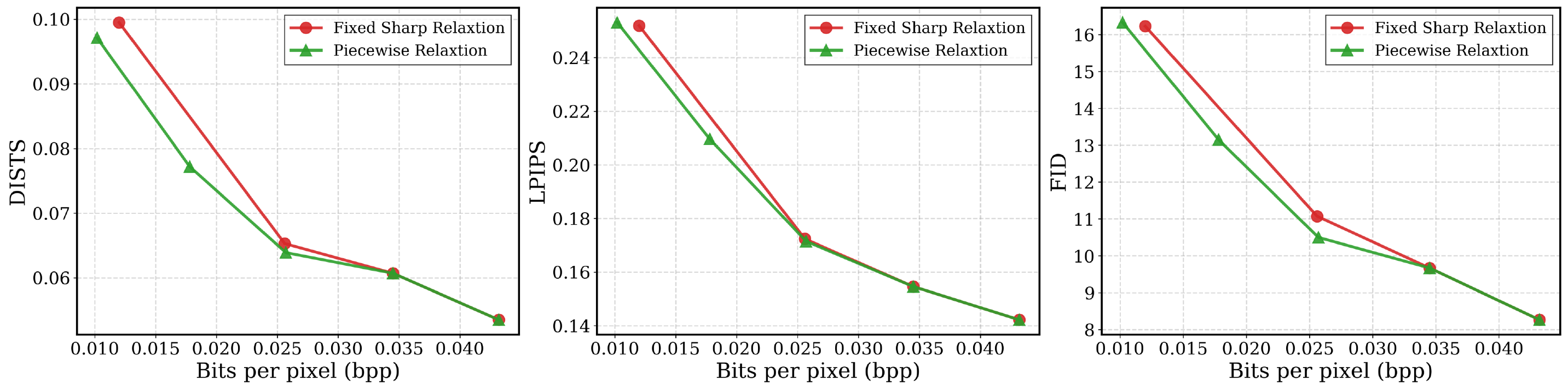} % 或设为0.9\columnwidth 等
  \vspace{-0.3cm}
  \caption{Effectiveness of piecewise strategy (evaluated on Div2K validation set).}
  \label{fig:temprature_piecewise}
  \vspace{-0.55cm}
\end{figure*}

\textbf{High-resolution fine-tuning.} Models trained solely on ImageNet $256\times256$ patches generalize poorly to high-resolution datasets (DIV2K, CLIC2020). To improve cross-resolution performance, we fine-tune models on OpenImage ($512\times512$, batch size 4, lr $1\!\times\!10^{-5}$, $4\!\times\!10^5$ iterations) followed by DF2K (random crops from $512$ to $2048$, batch size 1, lr $5\!\times\!10^{-6}$). This strategy enables the model to adapt to varying resolutions, improving reconstruction quality on high-resolution images. Comparative results on DIV2K show progressive improvements from ImageNet-only, OpenImage, to DF2K fine-tuning (see Fig.~\ref{fig:High_reso_adaption}).

All primary experiments use four NVIDIA RTX~4090 GPUs; DF2K high-resolution fine-tuning is performed on a single NVIDIA RTX~Pro~6000 GPU.

\paragraph{RD trade-off and temperature.}

The rate loss is defined as the cross-entropy between the relaxed index distribution $p_{\text{soft}}$ and the entropy model prediction $q_{\bm{\psi}}$, with the softmax temperature $\tau$ controlling distribution sharpness. Smaller $\tau$ produces sharper $p_{\text{soft}}$, resulting in concentrated gradients and stronger effective rate constraint for a given $\lambda$, whereas larger $\tau$ smooths $p_{\text{soft}}$, producing more diffuse gradients and weaker rate constraints. Thus, $\tau$ and $\lambda$ must be jointly considered to achieve the desired rate--distortion trade-off.

Following empirical observations, we choose $\tau$ and $\lambda$ based on the bitrate regime. In the ultra-low bitrate regime (bpp $<0.025$), smoother relaxation ($\tau=0.1$) tends to yield better RD performance, whereas in higher bpp regimes (bpp $>0.025$), sharper relaxation ($\tau=0.01$) generally performs better. Correspondingly, $\lambda$ is set to $\{4.8,7.2,12\}$ for low-bitrate models ($\tau=0.1$) and $\{0.8,1.2\}$ for higher-bitrate models ($\tau=0.01$) to maintain reasonable rate constraints.

The necessity of this piecewise strategy is confirmed by comparing it to a fixed-sharp relaxation baseline ($\tau=0.01$, $\lambda=\{0.8,\,1.2,\,1.8,\,2.4\}$). As shown in Fig.~\ref{fig:temprature_piecewise}, the piecewise strategy improves RD performance at ultra-low bpp, highlighting the effectiveness of piecewise relaxation.

\section{Experiments}

\subsection{Evaluation of third-party models.}
We compare RDVQ against recent state-of-the-art open-source learned image compression methods, including DDCM~\cite{ddcm}, RDEIC~\cite{li2025rdeic}, ResULIC~\cite{resulic}, MS-ILLM~\cite{msillm}, PerCo~\cite{Perco}, OSCAR~\cite{guo2025oscar}, DLF~\cite{xue2025dlf}, and StableCodec~\cite{zhang2025stablecodec}, as well as the classical codec VVC~\cite{VVC}. For DDCM, RDEIC, MSILLM, PerCo, OSCAR, DLF, StableCodec, and VVC, we run the official implementations under the same evaluation protocol as our RDVQ. For ResULIC, we report the performance provided in the official release, since their evaluation is conducted on the same benchmark settings.

\subsection{Evaluation Details}
For FID evaluation on CLIC2020 and DIV2K, following HiFiC~\cite{HiFiC}, we compute FID on patches rather than full images. Specifically, each input image is divided into overlapping $256\times256$ patches with a stride of 128, and FID is computed over the distribution of these patches.

\subsection{Quantitative Results}
\begin{figure*}[htbp]     
  \centering
  \includegraphics[width=0.85\textwidth]{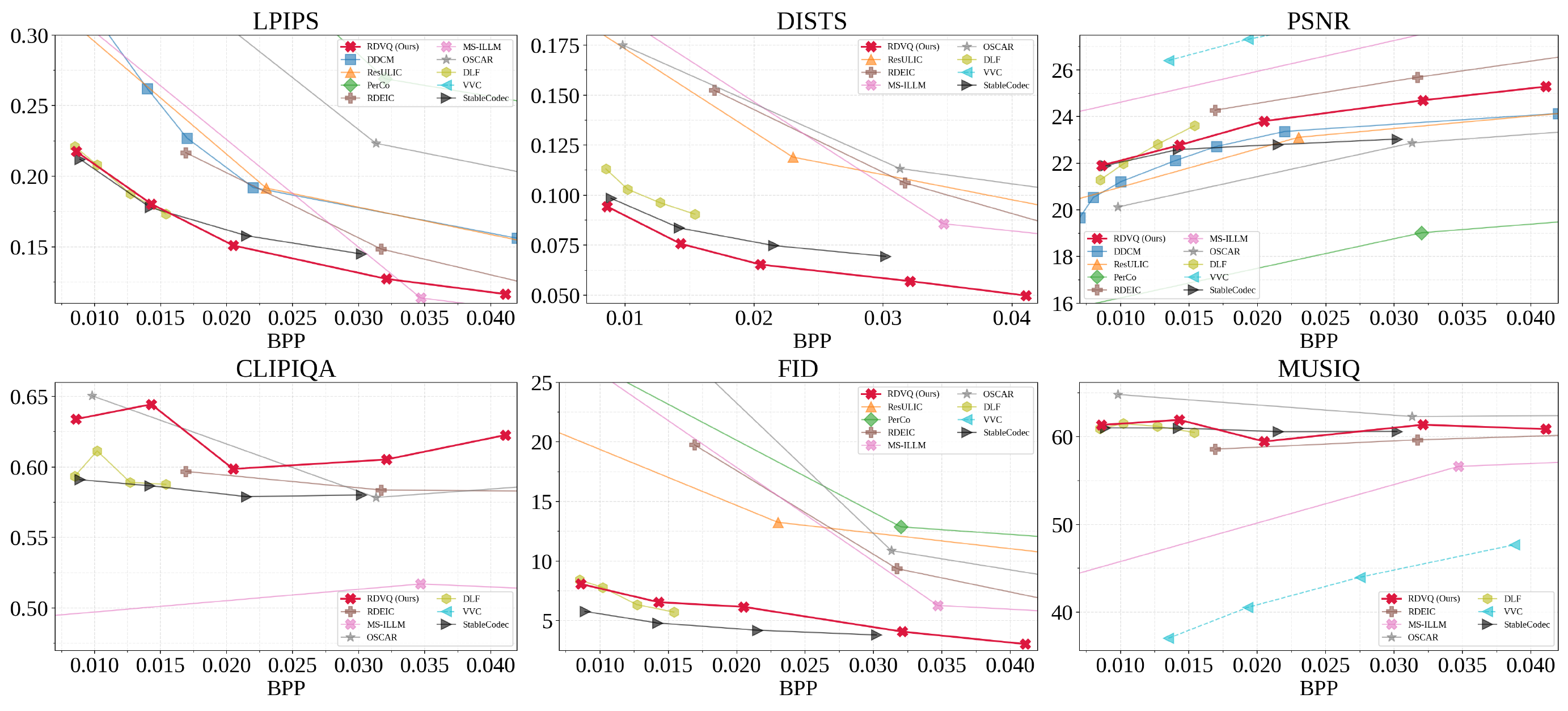} % 或设为0.9\columnwidth 等
  \caption{Rate-distortion curves on the CLIC 2020 testset.}
  % \vspace{-0.8cm}
  \label{Fid.RD_curve_cup_clic}
\end{figure*}

\begin{figure*}[htbp]     
  \centering
  \includegraphics[width=0.85\textwidth]{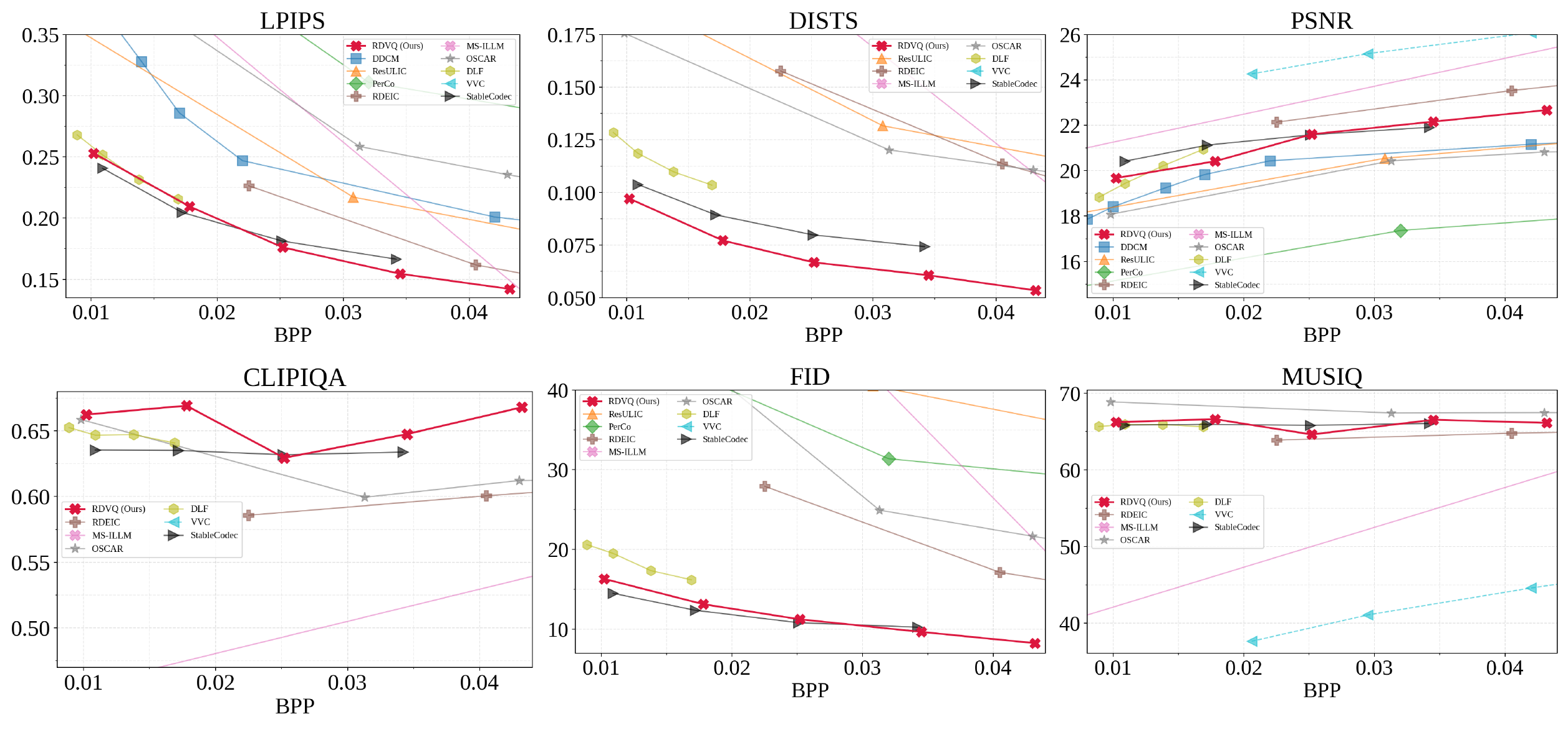} % 或设为0.9\columnwidth 等
  \caption{Rate-distortion curves on the DIV2K-val dataset.}
  % \vspace{-0.8cm}
  \label{Fid.RD_curve_cup_DIV2K}
\end{figure*}

\begin{figure*}[htbp]     
  \centering
  \includegraphics[width=0.52\textwidth]{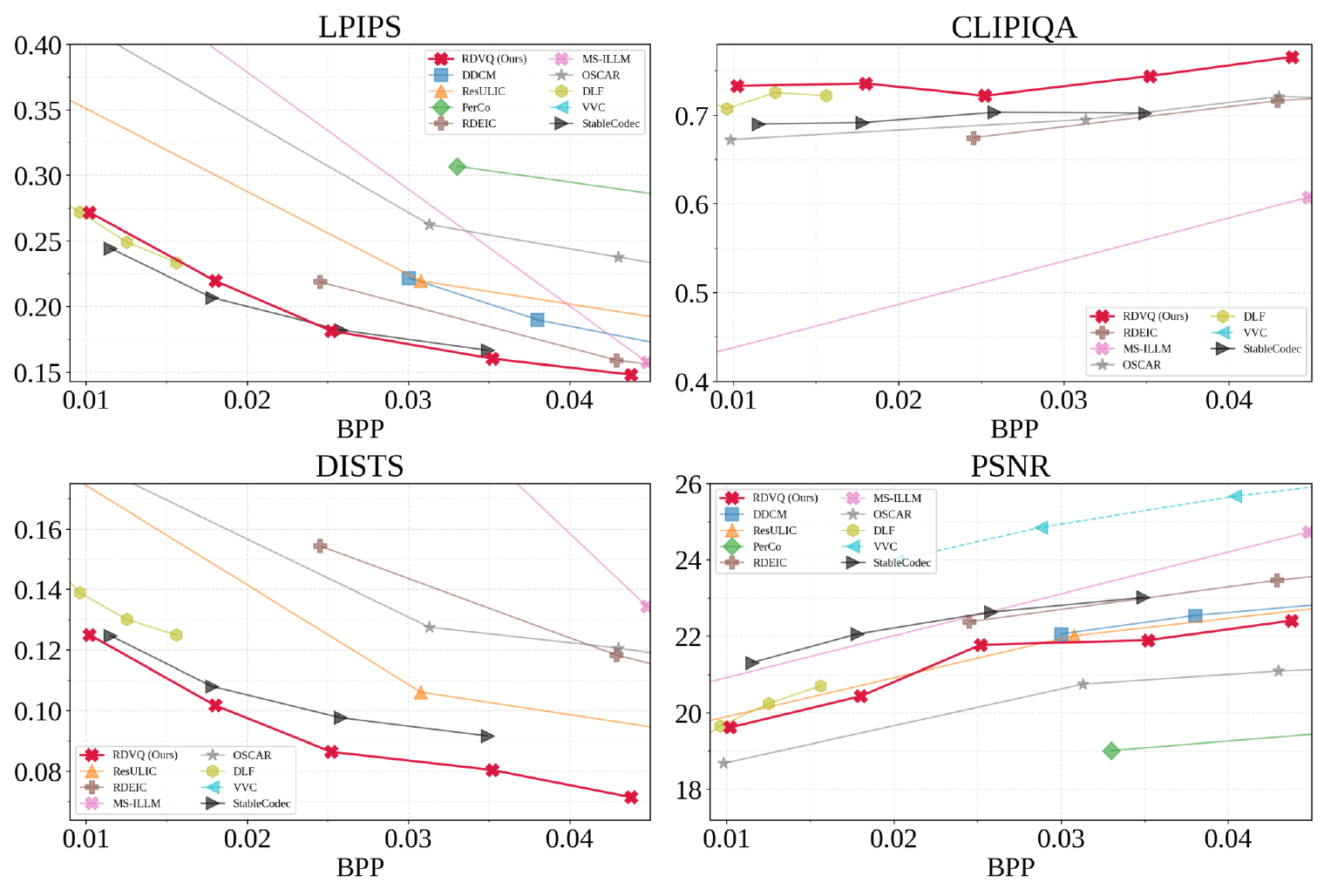} % 或设为0.9\columnwidth 等
  \caption{Rate-distortion curves on the Kodak dataset.}
  % \vspace{-0.8cm}
  \label{Fid.RD_curve_cup_Kodak}
\end{figure*}

Figure~\ref{Fid.RD_curve_cup_clic}, Fig.~\ref{Fid.RD_curve_cup_DIV2K}, and Fig.~\ref{Fid.RD_curve_cup_Kodak} present RD curves on CLIC2020-test, DIV2K-Val, and Kodak, respectively, using a pixel-level distortion metric. These plots provide a more detailed evaluation of RDVQ across different datasets.

\subsection{Test Time Rate Control Examples}
\begin{figure*}[htbp]     
  \centering
  \includegraphics[width=0.9\textwidth]{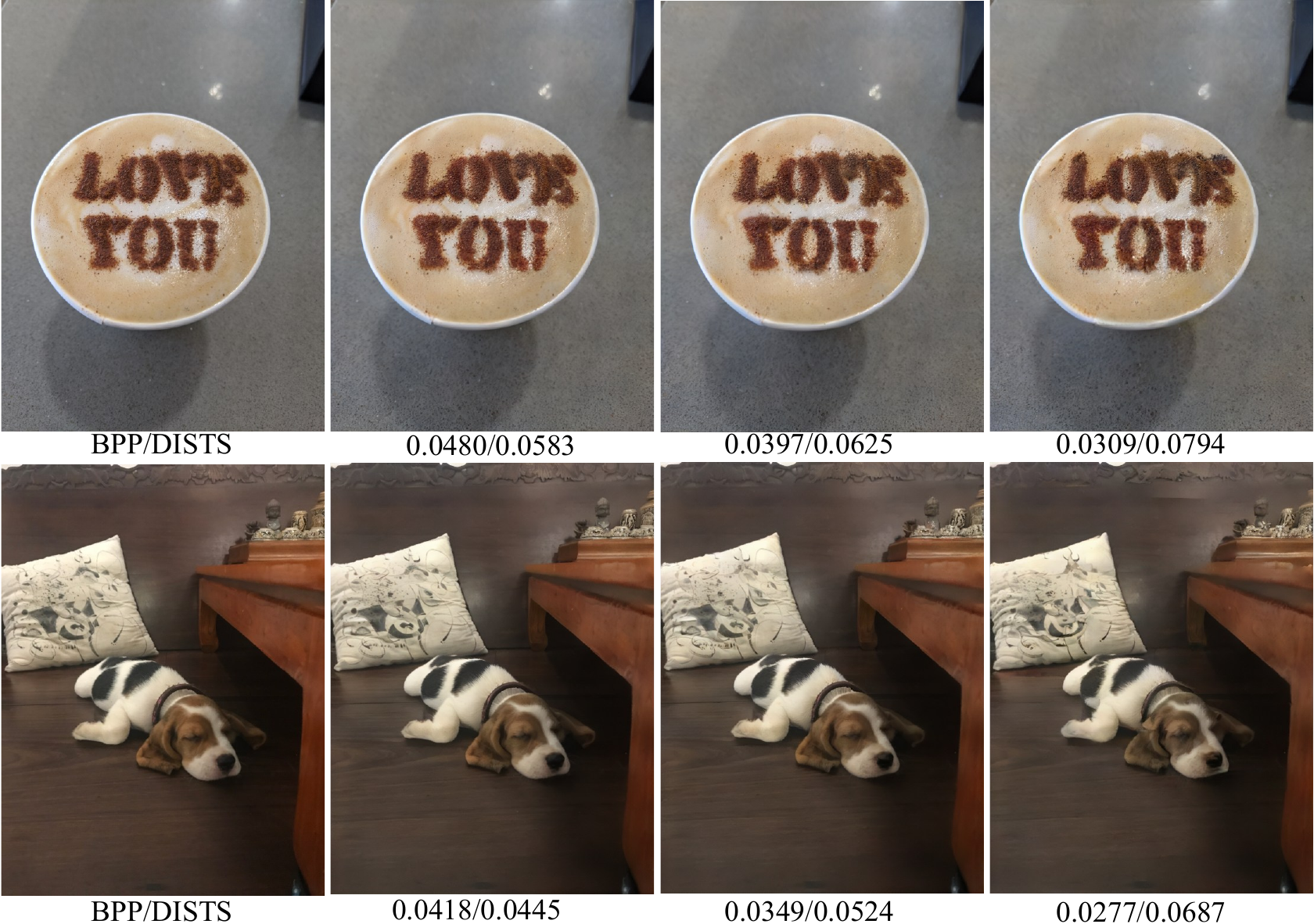} % 或设为0.9\columnwidth 等
  \caption{Test time rate control examples on CLIC2020 testset.}
  % \vspace{-0.8cm}
  \label{Fid.tta_CLIC}
\end{figure*}

\begin{figure*}[htbp]     
  \centering
  \includegraphics[width=0.9\textwidth]{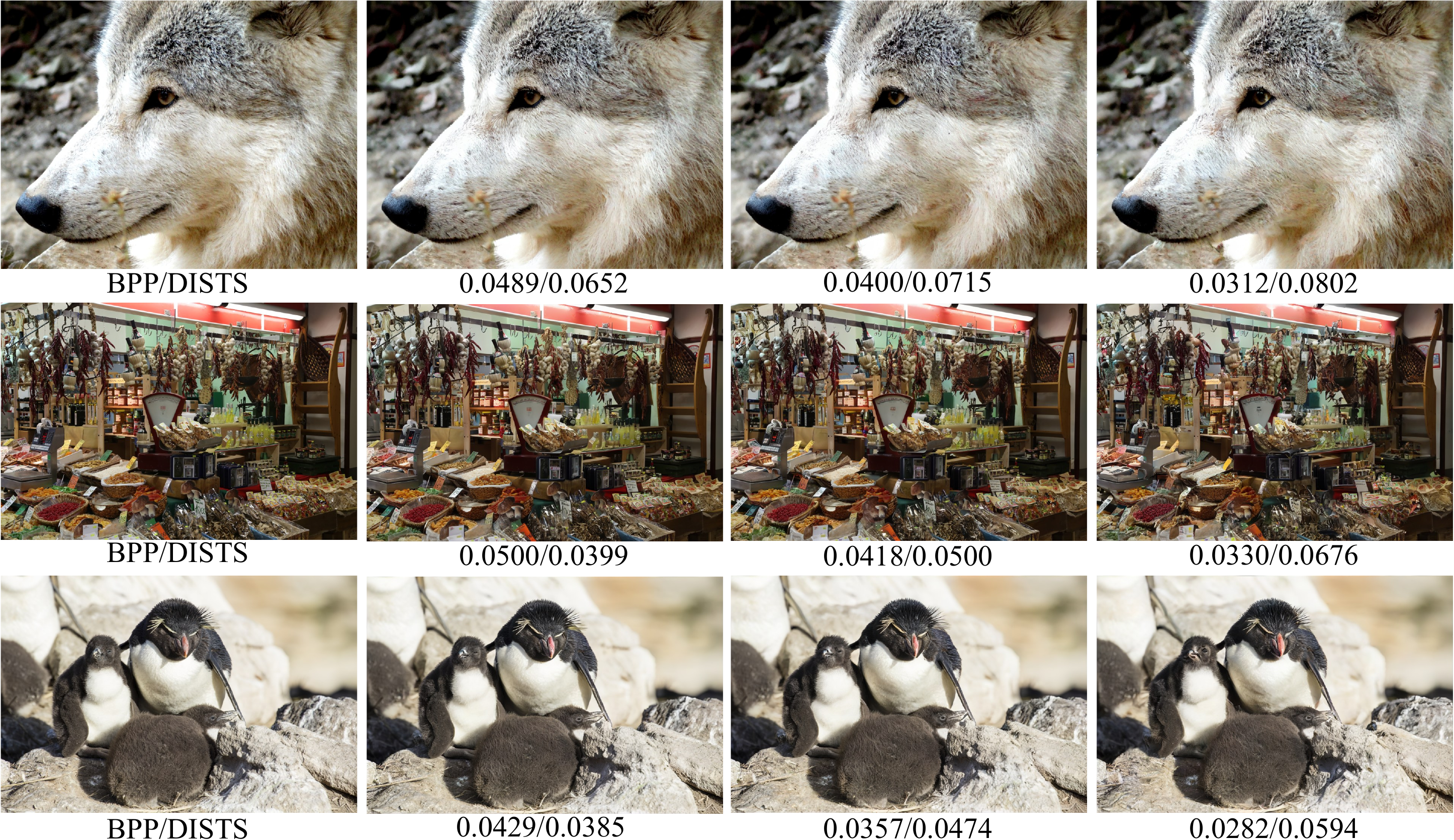} % 或设为0.9\columnwidth 等
  \caption{Test time rate control examples on DIV2K-val dataset.}
  % \vspace{-0.8cm}
  \label{Fid.tta_Div2k}
\end{figure*}

RDVQ enables test-time bitrate adjustment within a limited range via prefix transmission and autoregressive completion, without retraining. As shown in Fig.~\ref{Fid.tta_CLIC} and Fig.~\ref{Fid.tta_Div2k}, the model can smoothly vary the rate within 0.05–0.3 bpp on CLIC2020-test and DIV2K-Val, while maintaining high perceptual quality with no noticeable visual degradation.

\subsection{Additional Visual Examples}
\begin{figure*}[htbp]     
  \centering
  \includegraphics[width=0.95\textwidth]{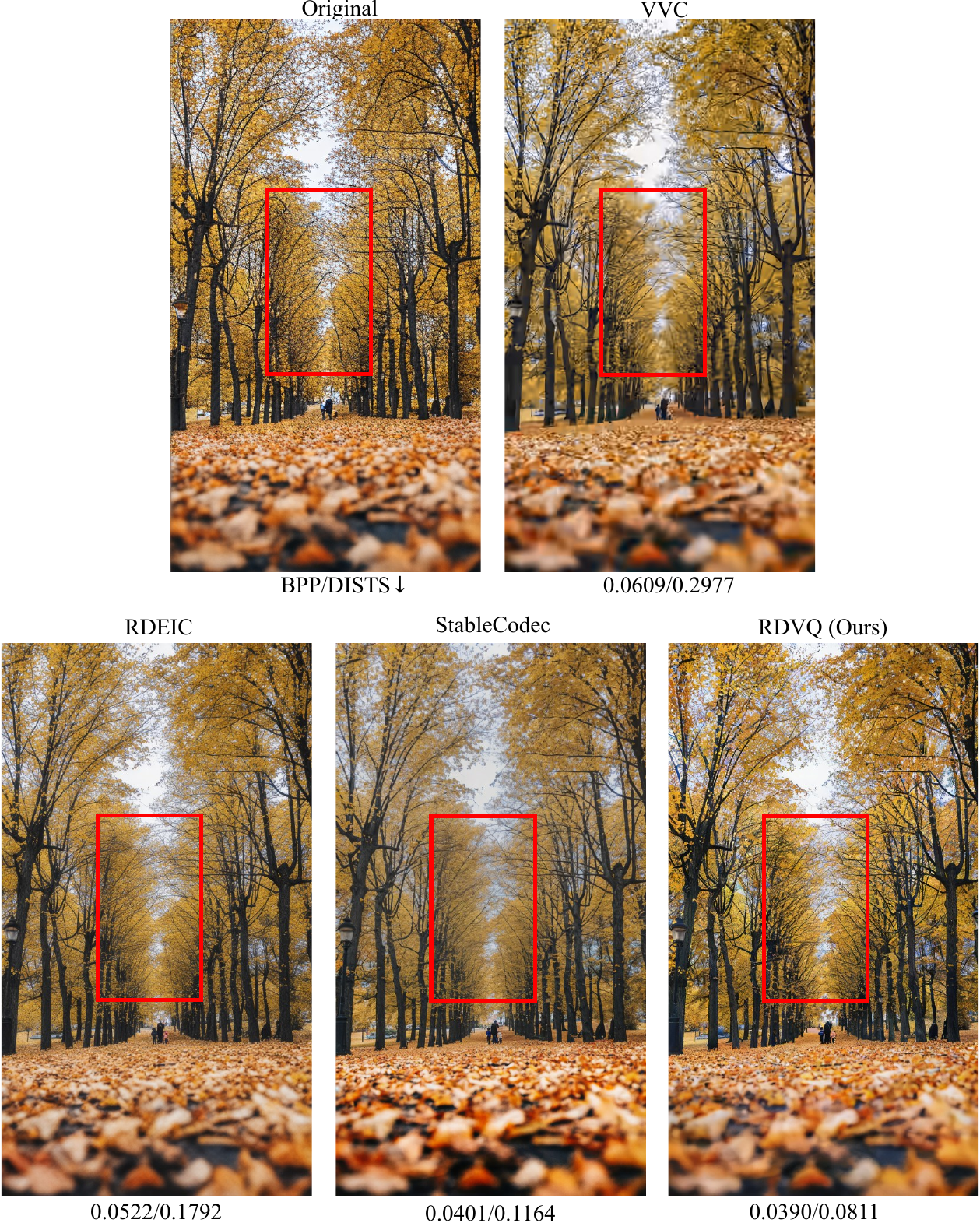} % 或设为0.9\columnwidth 等
  \caption{Visual examples and comparisons on 2K-resolution images from CLIC2020-test dataset.}
  %\vspace{-0.8cm}
  \label{fig:sup1}
\end{figure*}

\begin{figure*}[htbp]     
  \centering
  \includegraphics[width=0.95\textwidth]{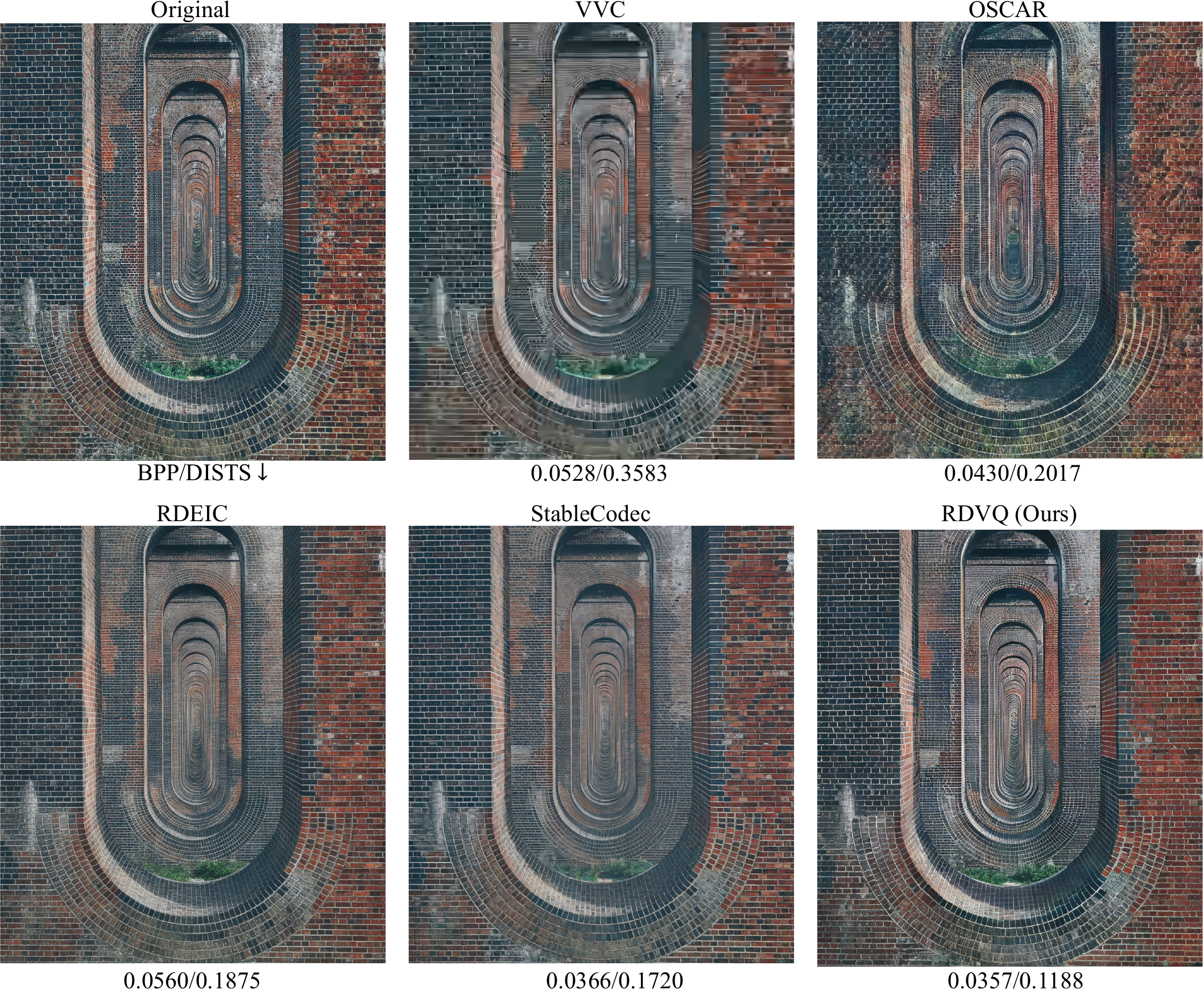} % 或设为0.9\columnwidth 等
  \caption{Visual examples and comparisons on 2K-resolution images from CLIC2020-test dataset.}
  %\vspace{-0.8cm}
  \label{fig:sup2}
\end{figure*}

\begin{figure*}[htbp]     
  \centering
  \includegraphics[width=0.95\textwidth]{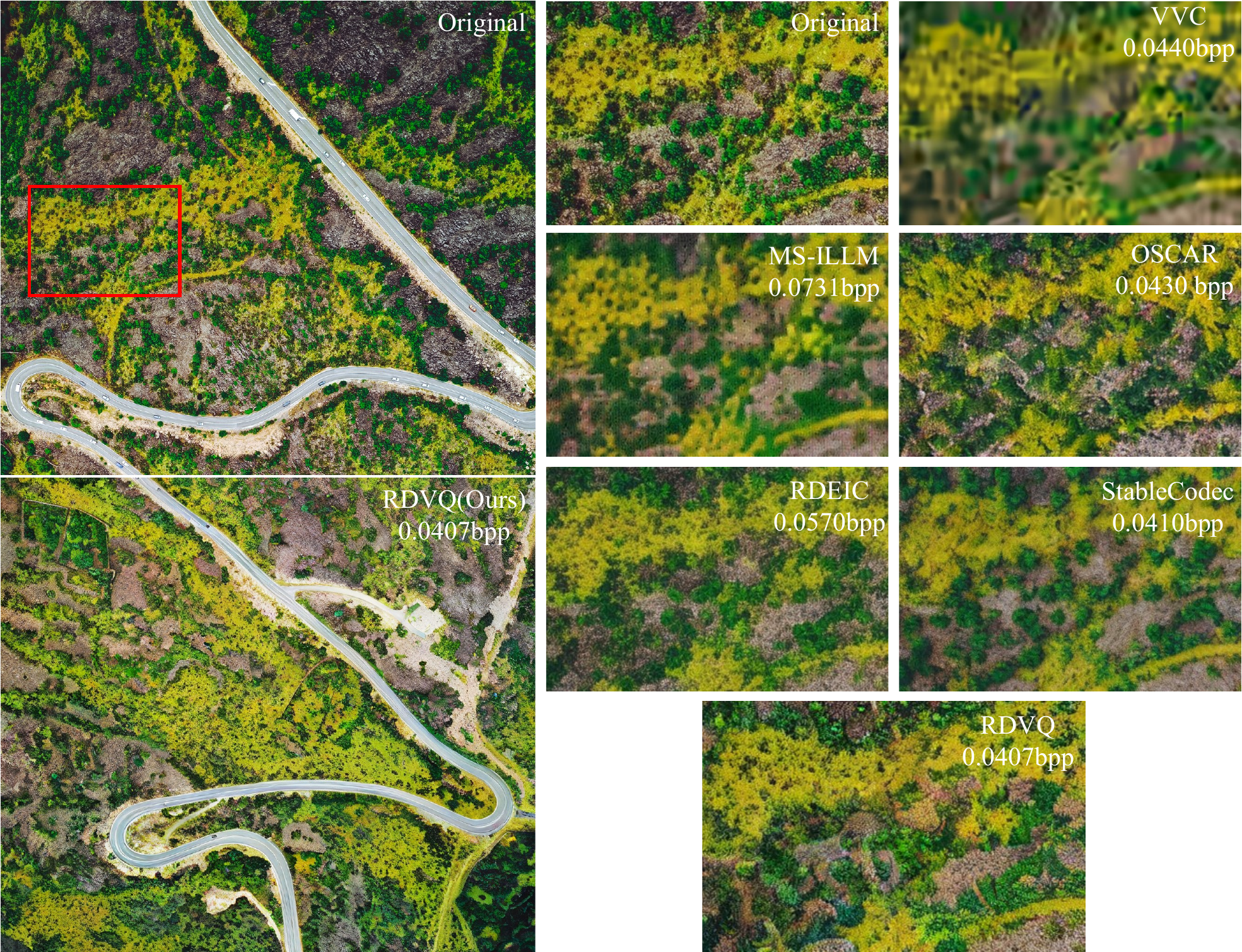} % 或设为0.9\columnwidth 等
  \caption{Visual examples and comparisons on 2K-resolution images from CLIC2020-test dataset.}
  %\vspace{-0.8cm}
  \label{fig:sup3}
\end{figure*}

\begin{figure*}[htbp]     
  \centering
  \includegraphics[width=0.95\textwidth]{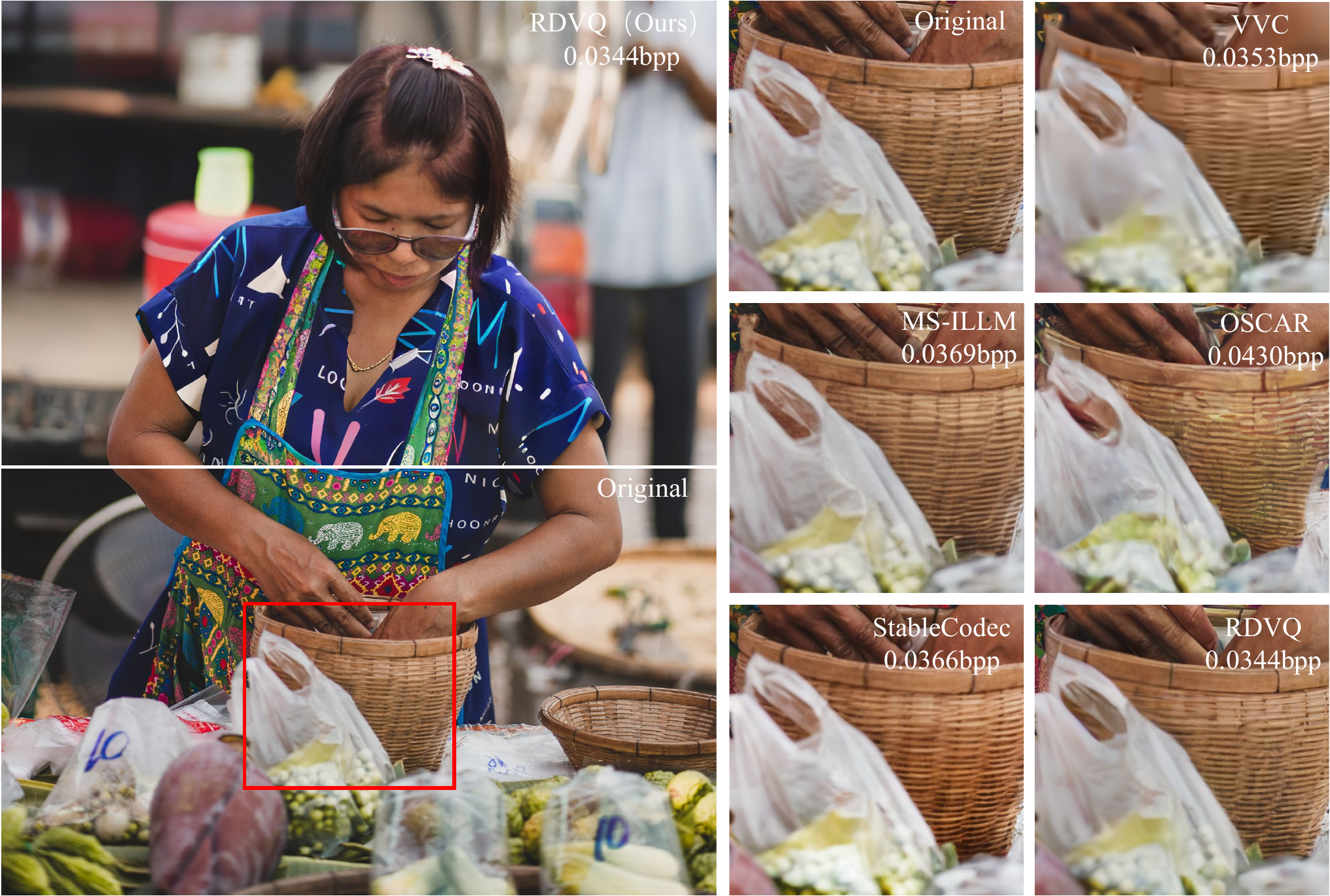} % 或设为0.9\columnwidth 等
  \caption{Visual examples and comparisons on 2K-resolution images from CLIC2020-test dataset.}
  %\vspace{-0.8cm}
  \label{fig:sup4}
\end{figure*}

\begin{figure*}[htbp]     
  \centering
  \includegraphics[width=0.95\textwidth]{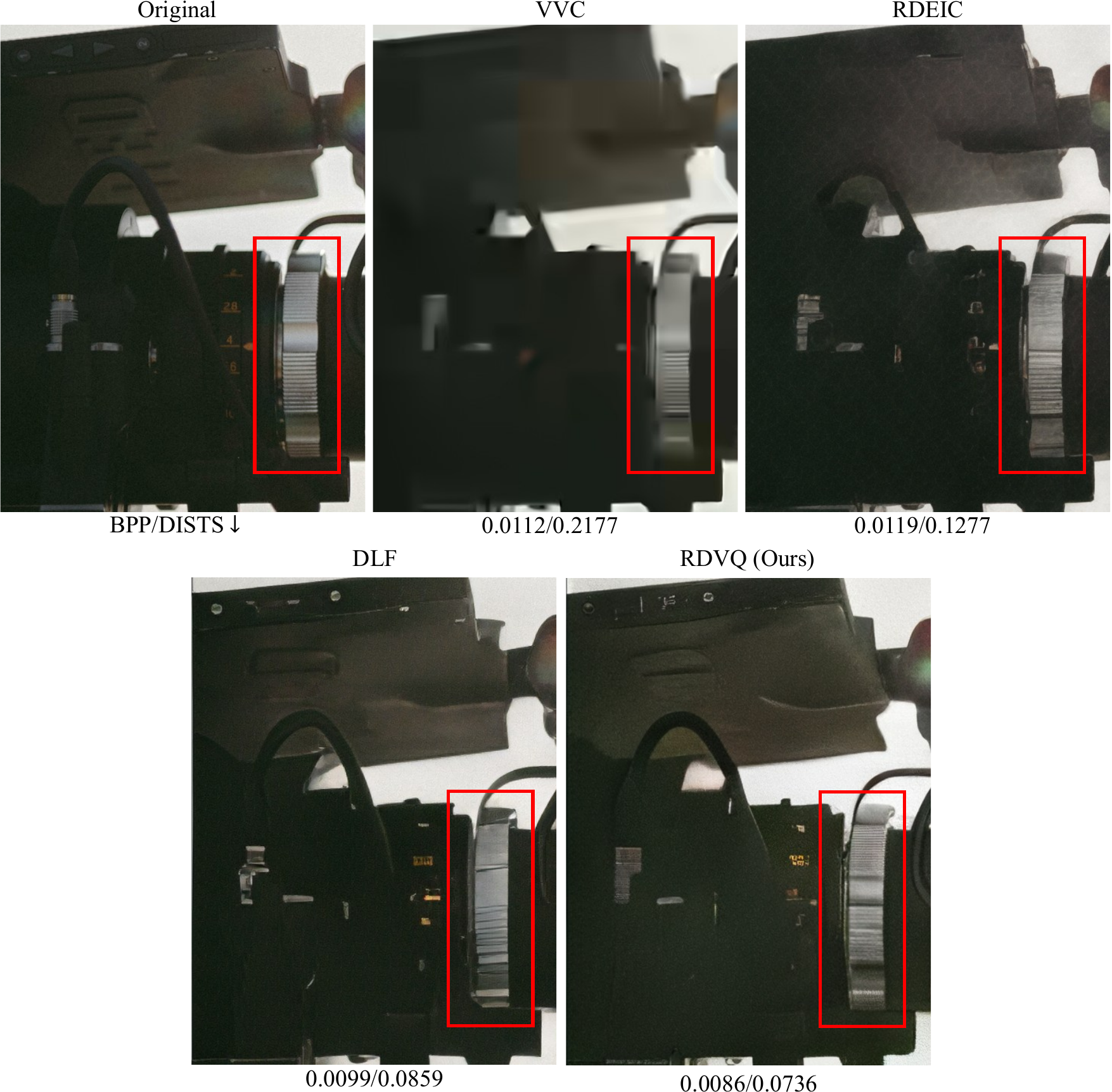} % 或设为0.9\columnwidth 等
  \caption{Visual examples and comparisons on 2K-resolution images from CLIC2020-test dataset.}
  %\vspace{-0.8cm}
  \label{fig:sup_low1}
\end{figure*}

\begin{figure*}[htbp]     
  \centering
  \includegraphics[width=0.95\textwidth]{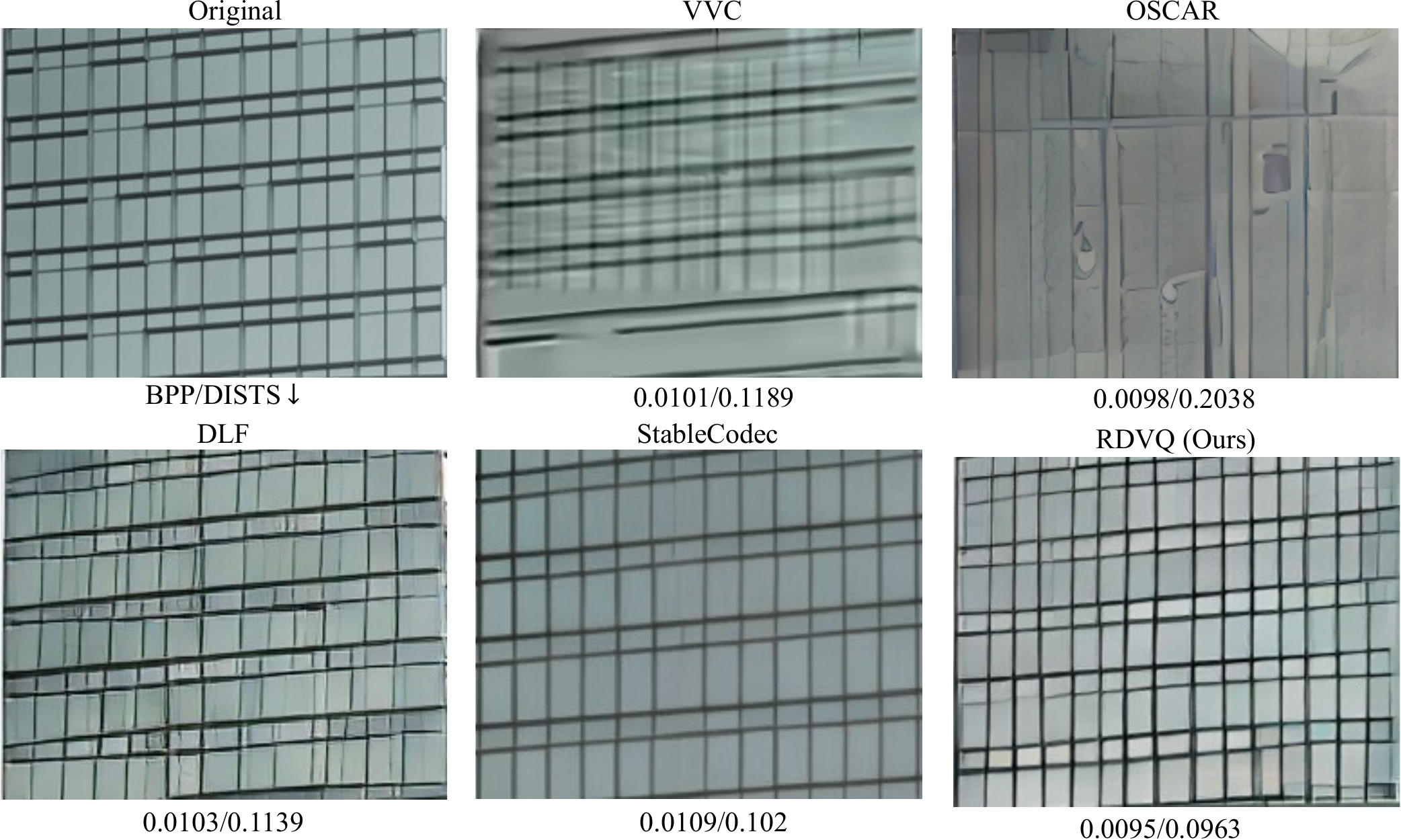} % 或设为0.9\columnwidth 等
  \caption{Visual examples and comparisons on 2K-resolution images from CLIC2020-test dataset.}
  %\vspace{-0.8cm}
  \label{fig:sup_low2}
\end{figure*}

\begin{figure*}[htbp]     
  \centering
  \includegraphics[width=0.95\textwidth]{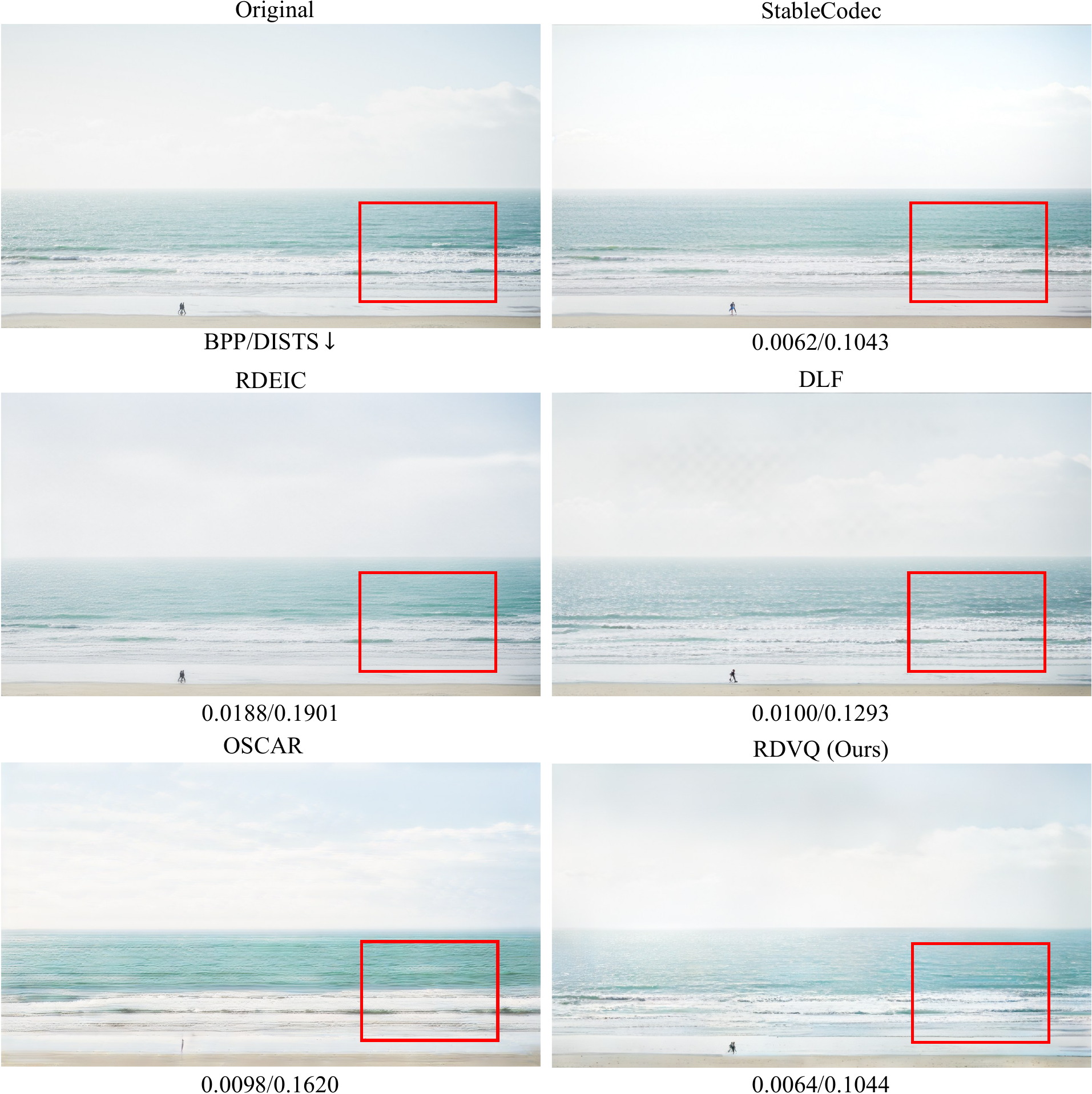} % 或设为0.9\columnwidth 等
  \caption{Visual examples and comparisons on 2K-resolution images from CLIC2020-test dataset.}
  %\vspace{-0.8cm}
  \label{fig:sup_low3}
\end{figure*}

We provide additional qualitative comparisons on CLIC2020-test in Figs.~\ref{fig:sup1}--\ref{fig:sup4} and Figs.~\ref{fig:sup_low1}--\ref{fig:sup_low3}. 
The former corresponds to relatively higher bitrates (bpp $=0.025\!\sim\!0.045$), while the latter shows more aggressive compression (bpp $=0\!\sim\!0.02$). 
In both cases, RDVQ consistently yields sharper details and better structural fidelity than competing methods, supporting the quantitative results.

\subsection{Comparison with other VQ-based methods}
We further compare our method with several VQ-based approaches~\cite{li2025onceforall, park2025diffo, xue2025one} on Kodak. 
As shown in Tab.~\ref{Tab:sup_VQ_baseline}, our method consistently outperforms all baselines in BD-DISTS, demonstrating superior rate--distortion efficiency.

\begin{table}[htbp]
\centering
\vspace{-2pt}
\footnotesize
\setlength{\tabcolsep}{4pt}
\begin{tabular}{lcccc}
\hline
Method & Ours & CGIC~\cite{li2025onceforall} & Diffo~\cite{park2025diffo} & OneDC~\cite{xue2025one} \\
\hline
BD-DISTS (\%, Kodak) & \textbf{0.0} & +5422.8 & +207.11 & +18.81 \\
\hline
\end{tabular}
% \vspace{-2pt}
\caption{Comparison with additional VQ-based methods on the Kodak dataset in terms of BD-DISTS. Lower is better.}
\vspace{-0.4cm}
\label{Tab:sup_VQ_baseline}
\end{table}

\end{document}